\documentclass{article} % For LaTeX2e
\usepackage{iclr2025_conference,times}

\usepackage{amsmath,amsfonts,bm}

\def\eqref#1{equation~\ref{#1}}
\def\1{\bm{1}}

\DeclareMathAlphabet{\mathsfit}{\encodingdefault}{\sfdefault}{m}{sl}
\SetMathAlphabet{\mathsfit}{bold}{\encodingdefault}{\sfdefault}{bx}{n}

\usepackage{minitoc}

\usepackage{hyperref}
\usepackage{url}
\usepackage{graphicx}
\usepackage{booktabs}
\usepackage{multirow}
\usepackage{bm}
\usepackage{xspace}
\newcommand{\nameshort}{\emph{LV}-Eval\xspace}
\title{\nameshort: A Balanced Long-Context Benchmark with 5 Length Levels Up to 256K}

\author{
  Tao Yuan$^2$, Xuefei Ning$^{1,\dagger}$, Dong Zhou$^2$, Zhijie Yang$^2$, Shiyao Li$^{1}$, Minghui Zhuang$^2$,
  \\ \textbf{ Zheyue Tan$^2$, Zhuyu Yao$^2$,  Dahua Lin$^{3, 4}$, Boxun Li$^{2,}$\thanks{Project Lead} , Guohao Dai$^{2, 5,\dagger}$, Shengen Yan$^{1,2}$,} \\ \textbf{ Yu Wang$^{1,}$}\thanks{Correspondence to Yu Wang \texttt{<yu-wang@tsinghua.edu.cn>}, Xuefei Ning \texttt{<foxdoraame@gmail.com>}, Guohao Dai \texttt{<daiguohao@sjtu.edu.cn, daiguohao@infini-ai.com>}.} \\
$^1$ Tsinghua University, $^2$ Infinigence-AI, $^3$ Shanghai Artificial Intelligence Laboratory \\
$^4$ The Chinese University of Hong Kong, $^5$ Shanghai Jiao Tong University
}

\iclrfinalcopy % Uncomment for camera-ready version, but NOT for submission.
\begin{document}

\doparttoc % Tell to minitoc to generate a toc for the parts
\faketableofcontents % Run a fake tableofcontents command for the partocs

\maketitle

\begin{abstract}
  State-of-the-art large language models (LLMs) are now claiming remarkable supported context lengths of $256k$ or even more. In contrast, the average context lengths of mainstream benchmarks are insufficient ($5k$-$21k$), and they suffer from potential knowledge leakage and inaccurate metrics, resulting in biased evaluation. This paper introduces \nameshort, a challenging long-context benchmark with five length levels ($16k$, $32k$, $64k$, $128k$, and $256k$) reaching up to $256k$ words.
\nameshort features two main tasks, single-hop QA and multi-hop QA, comprising 11 bilingual datasets. The design of \nameshort has incorporated three key techniques, namely confusing facts insertion (CFI), keyword and phrase replacement (KPR), and keyword-recall-based metric design. 
The advantages of \nameshort include controllable evaluation across context lengths, challenging test instances with confusing facts, mitigated knowledge leakage, and more objective evaluation. 
We evaluate 15 LLMs on \nameshort and conduct ablation studies on the benchmarking techniques. The results reveal that:
(i) Moonshot-v1 and recent large-scale open-source models, such as Qwen-2.5-72B and Llama-3.1-70B, achieve the highest performance on \nameshort, particularly at lengths below $64k$.
(ii) Models exhibit distinct score trends. For example, GLM-4-9B-128k, Yi-6B-200k, and Llama3-8B-1M exhibit a relatively gentle degradation of performance, but their absolute performances may not necessarily be higher than those of LLMs with shorter context lengths. (iii) LLMs' performances can significantly degrade in the presence of confusing information, especially in the pressure test of ``needle in a haystack''. (iv) Issues related to knowledge leakage and inaccurate metrics introduce bias in evaluation, and these concerns are alleviated in \nameshort. All datasets and evaluation codes are released at: \url{https://github.com/infinigence/LVEval}.
\end{abstract}

\section{Introduction}

% 先说两句llm兴起。再说长文本是llm一个重要特性。再说下面这些。而且要举出一些模型例子，以及ref。% such as machine dialog, text translation, document summarization, and novel creation. 
Large language models~(LLMs) have demonstrated exceptional performance on a variety of natural language processing tasks. 
The ability of long-context understanding is crucial for LLMs to deal with tasks based on longer contexts, such as books, lengthy chat history, and so on. Recently, extensive efforts have been devoted in enlarging the supported context length (i.e., the number of tokens that the model can accept as input) of LLMs.
These efforts have pushed the supported context length of LLMs from $2k$ tokens to $32k$ tokens~\citep{touvron2023llama, zheng2023judging, bai2023qwenvl, zeng2023glm130b, li2023long}, and some models have achieved a remarkable context length of $128k$ and $200k$~\citep{peng2023yarn, openai2023gpt4, Yi2023}. 

% In addition to having a smoothly decreasing perplexity (PPL) curve, long context models should also possess good performance in the aforementioned real-world application. It is crucial to assess these long context models in a reasonable and accurate manner. 

In contrast to the rapid evolution of the models' supported context length, existing benchmarks have lagged behind. The average word count in current long-context benchmarks typically falls within the range of $32k$~\citep{bai2023longbench, li2023loogle, an2023leval, shaham2023zeroscrolls, dong2023bamboo}, considerably shorter compared to the supported context lengths of state-of-the-art long-context models.
Moreover, previous benchmarks primarily consist of \textit{unaltered} public documents and articles. This could be problematic for two reasons: (i) the data might be involved in LLMs' training processes, and (ii) the facts within them might be common-sense facts found in other training resources. The presence of this issue, known as ``knowledge leakage''~\citep{zhou2023dont}, can lead to models answering questions with memorization or common-sense knowledge instead of understanding long-range contexts.
Last but not least, the automatic metrics employed in most of the existing benchmarks are susceptible to the variations in answer format and the inclusion of irrelevant words. 
Such metrics struggle to accurately assess the answer quality.

\begin{table}[t]
\centering
\resizebox{\linewidth}{!}{
\begin{tabular}{ccccccc}
\toprule
Benchmark    & \#Datasets & Avg \#Words & Min/Max Words & Length Levels & Opt. Metric & Lang. \\\midrule
ZeroSCROLLS~\cite{shaham2023zeroscrolls} & 10          & 13,556       & 1,023/320,774          & none     &      & en       \\ 
LooGLE~\cite{li2023loogle}       & 7           & 21,247       & 10,927/246,182         & none     &      & en       \\
L-Eval~\cite{an2023leval}       & 20          & 12,993       & 2,119/170,256          & none      & \checkmark    & en       \\
BAMBOO~\cite{dong2023bamboo}       & 10           & 5,067       & 229/14,858          & $4k$,$16k$    &       & en+zh       \\
LongBench~\cite{bai2023longbench}    & 21           & 9,486        & 128/71,954            & $0$-$4k$,$4k$-$8k$,$8k$+ &   & en+zh    \\\midrule
\nameshort       & 11           & 102,380      & 11,896/387,406         & $16k$,$32k$,$64k$,$128k$,$256k$ & \checkmark & en+zh    \\ \bottomrule
\end{tabular}
}
\caption{Comparison of different long-context benchmarks. We count the number of words for the English datasets and the number of characters for the Chinese datasets. The punctuation marks are taken into account, while tabs, blank spaces, and newlines are not included.}
\label{tab:benchmark_comparison}
\vspace{-0.4cm}
\end{table}

To address these issues, we propose \nameshort, a bilingual benchmark with up to $256k$ words. \nameshort incorporates distractions and confusions to make the test more challenging, replaces keywords and rephrases sentences to prevent knowledge leakage, and employs a more accurate metric. 
We summarizes the key characteristics of \nameshort as follows:
\begin{itemize}
\item \textbf{Sufficiently long context length to evaluate state-of-the-art models}: \nameshort comprises 5 length levels with word counts of $16k$, $32k$, $64k$, $128k$, and $256k$. Test instances across these levels share the same set of question-answer (QA) pairs, and only differ in the context content and length. Testing on the same QA pairs with different context lengths facilitates a controllable evaluation of models' long-context ability.
% \item \textbf{Easy expansion for future needs of more context length:} We obtain long context data mainly through mixing up multiple articles, therefore the levels of length are easy to expanding for customized needs, and the synthetic long context challenges LLMs in attention to vital evidence while discarding irrelevant information.
\item \textbf{Incorporation of distraction and confusion to increase difficulty}: When constructing the context for each test instance, we mix up distracting documents and supporting documents. This approach evaluates the model's ability in pinpointing key information in a large bunch of distracting texts. In addition, we insert confusing facts generated by GPT-4 and revised by human annotators into the context. This assesses the model's capability to accurately reason in the presence of interference.
\item \textbf{Keyword and phrase replacement to mitigate knowledge leakage}: To mitigate the biased evaluation of long-context ability caused by knowledge leakage, we replace the keywords and phrases in the context and QA pairs. The replacement rules are annotated by human annotators. In this way, \nameshort requires LLMs to rely on the understanding of context to answer questions rather than relying on memorization or common-sense knowledge.
\item \textbf{Keyword-recall-based metric for more objective scoring}: Existing $N$-gram metrics such as the F1 score are sensitive to the format variations and non-informative words in the answer, which results in inaccurate scores. To address this, we manually annotate answer keywords and a blacklist of unrelated words. The golden answers are the critical words or sentences extracted from original ground-truth (GT) answers, while the word blacklist contains common and non-informative words such as `the', `a', `of', and so on. The metric calculation follows a two-stage procedure: the first stage calculates the recall of golden answer keywords. if the recall exceeds a certain threshold, the second stage will remove all the blacklisted words and then calculate the F1 score between the prediction and the GT answer. This metric design can get scores with higher objectivity.
\end{itemize}

% 下面先画图再改
\paragraph{Findings.} We evaluate 15 LLMs on \nameshort and summarize the main findings as follows:
% (i) Commercial LLMs generally outperform open-source LLMs when evaluated within length levels shorter than their claimed context length. However, their overall performance is surpassed by open-source LLMs with longer context lengths.
(i) Moonshot-v1 and recent large-scale open-source models, such as Qwen-2.5-72B and Llama-3.1-70B, achieve the highest performance on \nameshort, particularly at lengths below $64k$.
(ii) Models exhibit distinct score trends. For example, GLM-4-9B-128k, Yi-6B-200k, and Llama3-8B-1M exhibit a relatively gentle degradation of performance, but their absolute performances may not necessarily be higher than those of LLMs with shorter context lengths. (iii) LLMs' performances can significantly degrade in the presence of confusing information, especially in the pressure test of ``needle in a haystack''. (iv) Issues related to knowledge leakage and inaccurate metrics introduce bias in evaluation, and these concerns are alleviated in \nameshort.

% (i) Commercial LLMs generally outperform open-source LLMs when evaluated within length levels shorter than their claimed context length. However, their overall performance is surpassed by open-source LLMs with longer context lengths. 
% (ii) Extremely long-context LLMs, such as Yi-6B-200k, exhibit a relatively gentle degradation of performance, but their absolute performances may not necessarily be higher than those of LLMs with shorter context lengths. 
% (iii) LLMs' performances can significantly degrade in the presence of confusing information, especially in the pressure test of ``needle in a haystack''. 
% (iv) LLMs perform better on the single-hop QA task compared to the multi-hop QA task.

\begin{table}[t]
\centering
\resizebox{\columnwidth}{!}{%
\begin{tabular}{c|cclcccc}
\toprule
Task &
  Dataset &
  CFI  &
  \#KPR  &
  AK  &
  Language &
  \#QA pairs &
  \#Contexts \\ \midrule
\multirow{6}{*}{Single-hop QA}   & \textbf{lic-mixup}              & \checkmark         &                         & \checkmark & zh & 197 & 985          \\
                                & \textbf{loogle-SD-mixup}       & \multirow{2}{*}{} &                         & \checkmark & en & 160 & 800          \\
                               & \textbf{cmrc-mixup}             &                   & \multicolumn{1}{c}{786} &           & zh & 200 & 1,000         \\
                               & \textbf{multifieldqa-en-mixup} & \checkmark         & \multicolumn{1}{c}{476} & \checkmark & en & 101 & 505          \\
                               & \textbf{multifieldqa-zh-mixup} & \checkmark         & \multicolumn{1}{c}{424} & \checkmark & zh & 133 & 665          \\
                               & \textbf{factrecall-en}          & \checkmark         & \multicolumn{1}{c}{3}   & \checkmark & en & 1   & 200$\times$5 \\
                               & \textbf{factrecall-zh}          & \checkmark         & \multicolumn{1}{c}{3}   & \checkmark & zh & 1   & 200$\times$5 \\ \midrule
\multirow{5}{*}{Multi-hop QA}  & \textbf{dureader-mixup}         & \multirow{3}{*}{} & \multirow{3}{*}{}       &           & zh & 176 & 880          \\
                               & \textbf{loogle-CR-mixup}       &                   &                         & \checkmark & en & 99  & 495          \\
                               & \textbf{loogle-MR-mixup}       &                   &                         & \checkmark & en & 139 & 695          \\
                               & \textbf{hotpotwikiqa-mixup}     & \checkmark         & \multicolumn{1}{c}{232} & \checkmark & en & 124 & 620          \\ \bottomrule
\end{tabular}%
}
\caption{Data statistics of \nameshort. The abbreviations ``CFI'', ``KPR'', ``AK'' stand for ``Confusing Fact Insertion'', ``Keyword and Phrase Replacement'', and ``Answer Keywords'', respectively. ``\#KPR'' is the number of KPR rules. Note that in \textbf{factrecall-en} and \textbf{factrecall-zh}, all QA pairs are the same across all test instances, i.e., there is only one unique QA pair for each of the two datasets.}
\label{tab:statistics}
\vspace{-0.5cm}
\end{table}

\section{Related Work}
\label{sec:rw}

\paragraph{Long-Context Benchmarks.}
Table~\ref{tab:benchmark_comparison} provides a summary of existing long-context benchmarks, including ZeroScrolls~\citep{shaham2023zeroscrolls}, LooGLE~\citep{li2023loogle}, L-Eval~\citep{an2023leval}, BAMBOO~\citep{dong2023bamboo}, and LongBench~\citep{bai2023longbench}.
ZeroScrolls, LooGLE, and L-Eval are monolingual benchmarks without explicit length level partition. Their average word counts are $\sim$$14k$, $\sim$$21k$ and $\sim$$13.5k$, respectively. In order to evaluate the model's capability across various context lengths, BAMBOO and LongBench have designed various length levels. However, the word counts ($\sim$$5k$, $\sim$$9.5k$) of the contexts in these two benchmarks are notably smaller than the supported context length of state-of-the-art long-context models, making them unsuitable for evaluating the claimed extremely long-context understanding ability. In contrast, \nameshort contains five length levels, up to $256k$ words, each with the same set of QA pairs for controllable evaluation.

In terms of metric design, L-Eval introduces a length-instruction-enhanced metric to mitigate the undesired impact of the answer length on metric scores. Additionally, L-Eval proposes to use LLMs to assist in scoring. In \nameshort, we ask human annotators to mark the answer keywords and create a non-informative word blacklist, and propose a two-stage metric to focus more on the answer keywords while reducing the influences of non-informative words.

The ``needle in a haystack'' (NIAH) task~\citep{needle2023} has been very popular in assessing the long-context retrieval ability of LLMs. A recent benchmark, RULER~\citep{hsieh2024ruler}, identifies the necessity of extending the NIAH task. It extends NIAH with diverse types and quantities of needles and distracting information. In our work, \nameshort also includes an extended version of the NIAH task with our confusing fact insertion and keyword-and-phrase replacement techniques.
%Regarding other benchmarking techniques, a contemporary work~\citep{hsieh2024ruler} proposes including distracting information in the context of a "needle in a haystack" task, which shares a similar spirit with our context mixing technique.

\paragraph{Long-Context Techniques.} Considerable efforts have been devoted to enhancing the long-context abilities of LLMs. One line of work focuses on making LLMs have extended context sizes without fine-tuning and behave normally on inputs longer than their training context lengths. The design and extrapolation method of the position encoding module~\citep{su2024roformer,press2021train,ntkaware} is crucial for this goal. Besides, several sparse attention techniques~\citep{han2023lm,streamingllm} have also been proposed to avoid model collapse. These sparse attention techniques also alleviate the quadratic complexity w.r.t. the sequence length.

There are many other strategies aimed at enabling LLMs to effectively leverage long input contexts. The most commonly utilized strategy is long-context fine-tuning~\citep{xiong2023effective,li2023long,yarn}. 
For instance, YaRN~\citep{yarn} conducts fine-tuning with $64k$ and $128k$ context lengths starting with Llama2-7B/13B, and Yi-6B-200k~\citep{Yi2023} is trained with $200k$ context length starting with its $4k$ variant.
Other strategies include the recurrent- or memory-based architecture~\citep{dai2019transformer,wu2021memorizing,martins2022former,zhou2023recurrentgpt,liang2023unleashing}, and the retrieval- or summarization-based context compression techniques~\citep{khandelwal2019generalization,borgeaud2022improving,bai2023longbench,zhou2023recurrentgpt}, and so on. 

In this work, we evaluate LLMs of diverse context sizes, ranging from $4k$ to $200k$, most of which have incorporated advanced position encoding design and undergone long-context fine-tuning.

\section{\nameshort Benchmark}
\label{sec:benchmark}

\begin{figure}[th]
\centering
\includegraphics[width=1\linewidth]{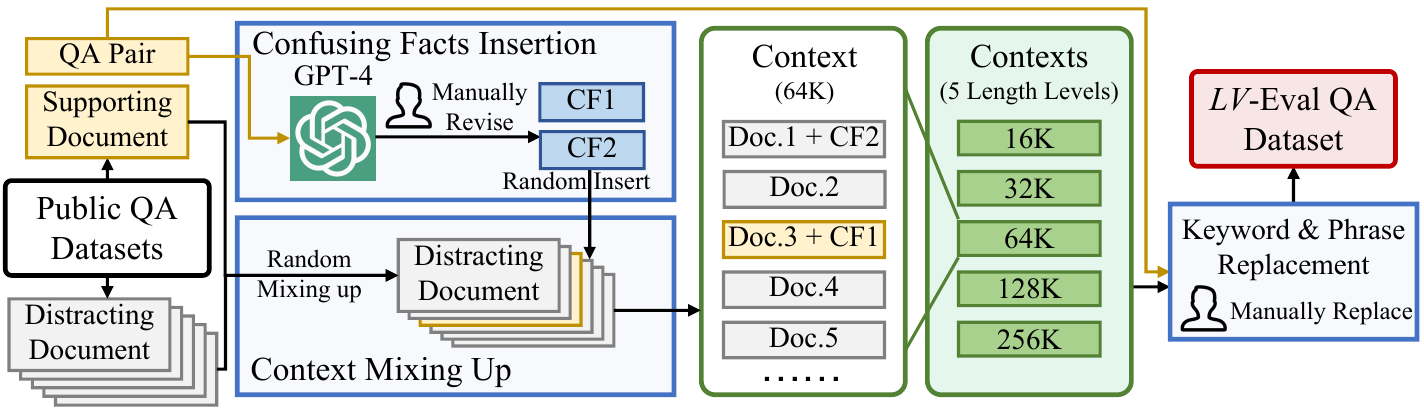}
%\vspace{-0.4cm}
\caption{The construction process of \nameshort. ``CF'' is short for ``Confusing Fact''.}
\label{fig:overall_construction}
%\vspace{-0.2cm}
\end{figure}

\nameshort focuses on two types of QA tasks: single-hop QA and multi-hop QA, and is comprised of 11 QA datasets (6 in English and 5 in Chinese). The data statistics for \nameshort are outlined in Table~\ref{tab:statistics}.
Each test instance in \nameshort comprises three parts: a context ($C$), a question ($Q$), and a GT answer ($A$), where $C$ is a synthetic document containing the information required to answer $Q$.

Datasets in \nameshort are constructed with existing public datasets as the source, except for factrecall-en and factrecall-zh, which are constructed using the data from \textit{PG19}~\citep{pg19} dataset and \textit{Journey to the West} book. Each dataset consists of five subsets of different lengths: $16k$, $32k$, $64k$, $128k$, and $256k$. All five subsets share the same question-answer (QA) pairs, meaning there are five contexts of varying lengths for each QA pair. This allows for a controllable evaluation of models' long-context ability when testing the same set of questions with different context lengths. In total, \nameshort comprises 1,729 QA pairs and 1,729$\times$5 = 8,645 synthetic contexts.

Figure~\ref{fig:overall_construction} illustrates the construction process of \nameshort. For \textbf{factrecall-en} and \textbf{factrecall-zh}, we write one QA pair for each dataset. For the rest 9 out of the 11 datasets, we first choose a specific number of QA pairs from existing QA datasets (Section~\ref{sec:data_source}). 
Then, for each unique QA pair, we go through three procedures to construct the context  (Section~\ref{sec:context_construction}):
\begin{enumerate}
\item \textbf{Context mixing up} (Section~\ref{sec:mixup}): We first construct five contexts of different lengths by mixing up supporting documents corresponding to the QA pair and several distracting documents. 
For \textbf{factrecall-en} and \textbf{factrecall-zh}, we mix the supporting evidence of the single QA pair with distracting documents from two books.
For other datasets, the distracting documents are unrelated to the question and are chosen from the context documents corresponding to non-selected QA pairs in the same source dataset.
\item \textbf{Confusing Facts Insertion (CFI)} (Section~\ref{sec:confusion}): Then, in some datasets, we introduce confusing facts by generating them with GPT-4, manually revising them, and randomly inserting these into the context. 
These confusing facts bear similarities to the original supporting facts but are factually different, without contradicting the original information.
This helps make the test instances more challenging.
\item \textbf{Keyword and Phrase Replacement (KPR)} (Section~\ref{sec:replacement}): Finally, to reduce the impacts of knowledge leakage on evaluation results, we manually replace some keywords and phrases in the context and the QA pairs.
\end{enumerate}

When evaluating the generated answer, to mitigate the bias in existing metrics, we manually annotate the keywords in the GT answer and adjust the metric to focus more on the keywords (Section~\ref{sec:optimized_metrics}).

During the benchmark construction, we employ and guide human annotators to revise the confusing facts, replace the keywords and phrases, and annotate the keywords in GT answers. The details of our annotation process are introduced in Appendix~\ref{sec:annotation}.

%\subsection{Dataset Collection and Cleaning}
\subsection{Data Source and QA Pair Construction}
\label{sec:data_source}
We construct 11 datasets (see Table~\ref{tab:statistics}) using public data sources, including \verb+Long-instruction-en2zh+~\citep{Long2023}, HotpotQA~\citep{yang2018hotpotqa}, 2WikiMultihopQA~\citep{ho2020constructing}, DuReader~\citep{tang2021dureaderrobust}, LooGLE~\citep{li2023loogle}, LongBench~\citep{bai2023longbench}, CMRC 2018~\citep{Cui_2019}, MultiFieldQA~\citep{bai2023longbench}, PG-19~\citep{pg19} and the book of \textit{Journey to the West}. The construction of QA pairs in each dataset is elaborated in Appendix~\ref{appendix:A}.
%Ten datasets are collected from existing public data sources, and two are self-constructed. According to the task type, they are divided into multi-hop QA, single-hop QA and summary.  ‘’Journey to the West‘’

\subsection{Context Construction}
\label{sec:context_construction}

%In this section, we highlight four key procedures in data construction of \nameshort. Concretely, we illustrate the method of Context mixing up, distracting facts insertion, Keyword and phrase replacement, and optimized metrics with answer keywords. Please refer to Appendix B for the complete annotation details and examples.

\begin{figure}[t]
\centering
\includegraphics[width=1\linewidth]{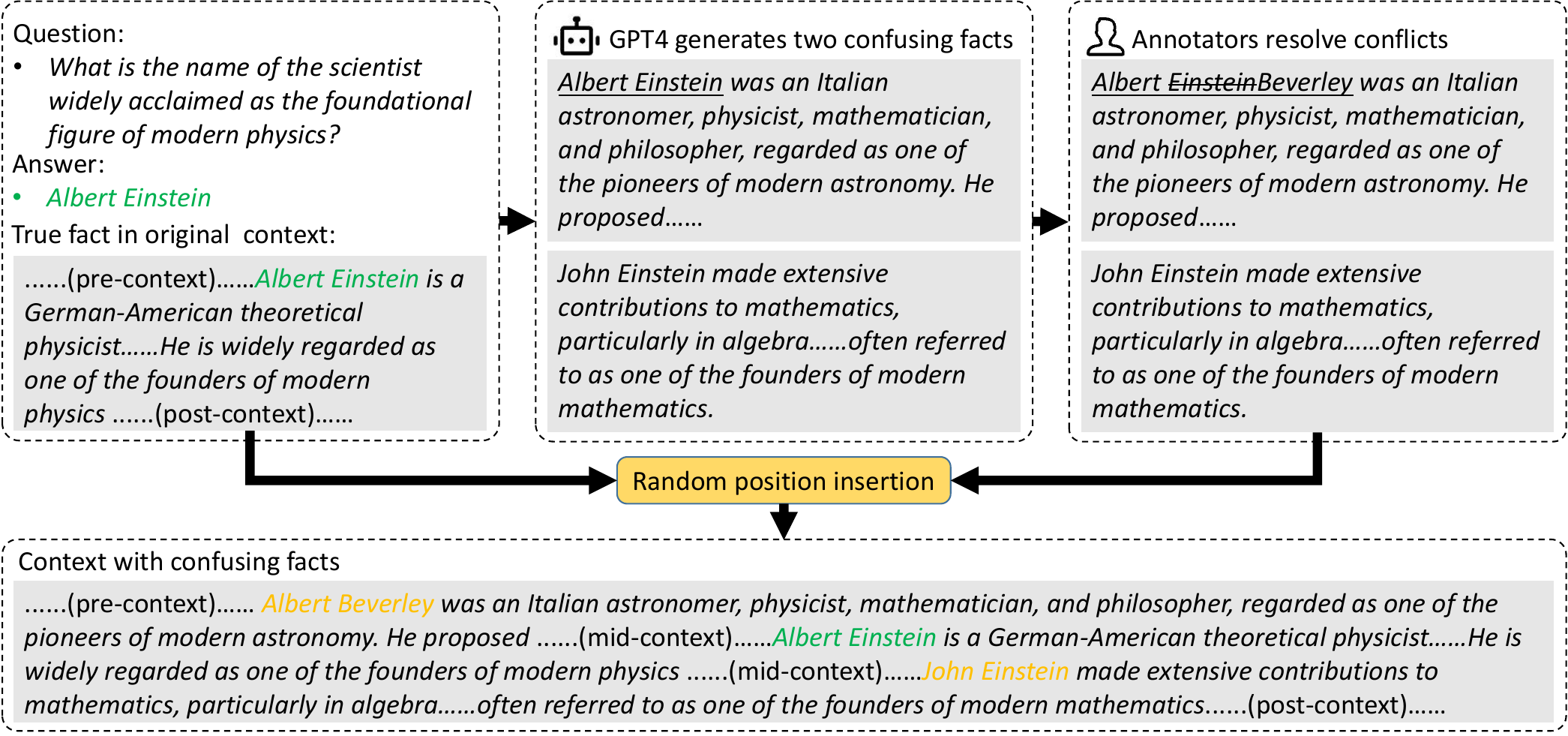}
%\vspace{-0.4cm}
\caption{Steps for CFI. Firstly, we prompt GPT-4 to generate two descriptions that are close to the original fact. Then we ask human annotators to resolve any conflicts in the generated facts. For example, the first generated confusing fact ``Albert Einstein was an Italian astronomer'' is in conflict with the original fact and the human annotator revise it to ``Albert Beverley was an Italian astronomer''. Finally, the confusing facts are inserted into a randomly position in the context.}
\label{fig:prc_confusing_fact_insert}
\end{figure}

\subsubsection{Context Mixing Up}
\label{sec:mixup}

%Apart from \textbf{Factrecall}, all of our QA datasets are constructed in the method of mixing up multiple documents. The two main objectives of the method are to extend the context length to meet the 5 length levels we need, and to add confusing information in the context to make evaluation more challenging. we filter out relatively longer documents from a certain data source to make the concatenated context less fragmented, and choose documents from the same source with the original article to avoid significant style difference. 

Can the LLMs identify the key evidences to answer the target question within a long context? To assess this ability, as shown in Figure~\ref{fig:overall_construction}, \nameshort randomly mixes the supporting documents with various distracting documents to generate five contexts of varying length for a given QA pair. For 9 out of the 11 datasets (excluding \textbf{factrecall-en} and \textbf{factrecall-zh}), the distracting documents are chosen from the contexts corresponding to the non-selected QA pairs in the source dataset. For \textbf{factrecall-en} and \textbf{factrecall-zh}, the distracting documents are extracted from the \textit{PG-19} dataset and the book of \textit{Journey to the West}. 

For each length level, we sample distracting documents one by one until the cumulative word count meets the desired length level. Then, we shuffle the supporting and distracting documents, prepend a string ``\verb|Passage i\n|'' to the $i$-th document, and concatenate them to form the final context. 

Note that in \textbf{hotpotwikiqa-mixup} and \textbf{dureader-mixup}, where  multiple supporting documents exist for each QA pair, instead of regarding the multiple supporting documents a single unit, we disperse and shuffle all supporting and distracting documents.

% \begin{figure}
% \centering
% \includegraphics[width=1\linewidth]{Figures/answer_keywords.pdf}
% \label{fig:ans_keywords_word_blacklist}
% \caption{Keyword-recall-based two-stage metric calculation. (a) The vanilla F1 score (red) is inflated high. With the keyword recall-based metric, the final score is set to zero due to the low recall of answer keywords. (b) The vanilla F1 score (red) is inflated high due to irrelevant words. By filtering of blacklisted words, the final score is better calibrated.}
% \end{figure}

% supporting distracting
\subsubsection{Confusing Facts Insertion}
\label{sec:confusion}

Can the LLMs identify the key evidences correctly if there are confusing facts in the context? To assess this ability, we apply CFI in \textbf{hotpotwikiqa-mixup}, \textbf{lic-mixup}, \textbf{multifieldqa-en-mixup}, \textbf{multifieldqa-zh-mixup}, \textbf{factrecall-en}, and \textbf{factrecall-zh}, which inserts similar, factually different, non-contradictory facts into the context. These facts might mislead less meticulous models, leading them to generate incorrect answers.

The generation process of the confusing facts goes as follows. Firstly, we use the question and answer as the input, and prompt GPT-4~\citep{openai2023gpt4} to generate two descriptions that are close to the original fact. The prompt for GPT-4 is shown in Figure~\ref{fig:confusing_fact_prompt}. Then, we ask human annotators to resolve any conflicts in the generated facts. As illustrated in Figure~\ref{fig:prc_confusing_fact_insert}, the generated confusing fact ``Albert Einstein was an Italian astronomer'' is in conflict with the original fact. Therefore, the human annotator revise it to ``Albert Beverley was an Italian astronomer''. After this generation and revising process, we insert the confusing facts into a randomly picked position between two sentences in the context. 

\begin{figure}[t]
\centering
\includegraphics[width=0.84\linewidth]{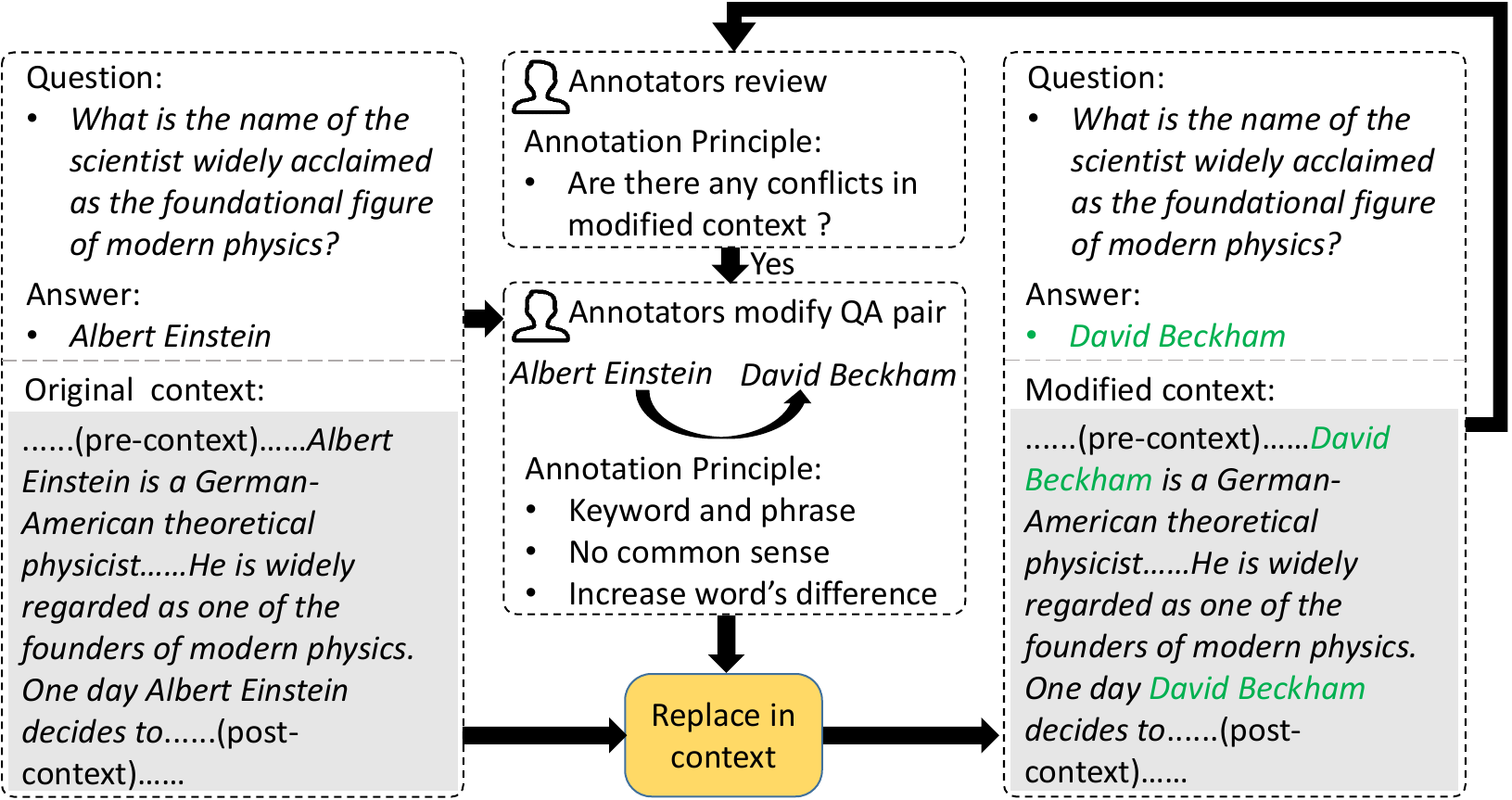}
\caption{Steps for KPR. First, given a QA pair, the annotators are asked to select keywords or phrases to replace and write a substitute for each. Then, the selected keywords and phrases are replaced throughout the context and QA pair. Finally, annotators will check
the modified context. If there is any conflict, the annotators are asked to revise the replacement rule until all conflicts are resolved.
}
\label{fig:prc_keyword_phrase_replace}
\end{figure}

\subsubsection{Keyword and Phrase Replacement}
\label{sec:replacement}

Knowledge leakage is an important concern in LLM evaluation~\citep{zhou2023dont}. On the one hand, the test data are usually collected from open-access sources, and we cannot fully rule out the possibility of their being involved in some LLMs' training process. On the other hand, some common-sense questions can be answered without referencing the provided context. Consequently, LLMs might rely on memorization and common-sense knowledge to answer the questions rather than fully understanding the context. This will cause inflated benchmark scores to overrate the long-context ability of models.

To mitigate the influences of knowledge leakage on the evaluation results, we conduct KPR according to manually crafted rules in %\textbf{dureader-mixup}, 
\textbf{hotpotwikiqa-mixup}, \textbf{cmrc-mixup}, \textbf{multifieldqa-en-mixup}, \textbf{multifieldqa-zh-mixup}, \textbf{factrecall-en}, and \textbf{factrecall-zh}. Specifically, given a QA pair, the annotators are asked to select keywords or phrases for replacement and write a substitute for each. %Concretely, 786, 232, 476, 424, 
%\todo{how many rules do we manually write for each dataset, can share some statistics. replace pairs: cmrc 786, hotpotwikiqa 232, mfqa-en 476, mfqa-zh 424, factrecall-en 3, factrecall-zh 3} 
After the selected keywords and phrases are replaced throughout the entire context, the annotators review the modified context to check and resolve any conflicts: If there are conflicts, the annotators are asked to revise the replacement rule until all conflicts are resolved. One example of the KPR process is shown in Figure~\ref{fig:prc_keyword_phrase_replace}. See Table~\ref{tab:statistics} for the statistics of the number of replacement rules.

\subsection{Metric Design}
\label{sec:optimized_metrics}

The quality evaluation of natural language generation is challenging. Current $N$-gram metrics, such as the F1 score, treat all words equally. The neglect of differences in word importance leads to evaluation bias. For example, in the sentence ``Attention is all you need'', the word ``attention'' carries the key information and is more important. However, the answer ``Attention matters'' will get a lower score than the answer ``CNN is all you need'', which is not what we expected. To this end, we adopt a two-stage metric calculation process. 

Specifically, to evaluate an answer $A'$, we first calculate the recall of several ``answer keywords'' in $A'$. When the recall exceeds a certain threshold (0.2 for Chinese dataset, 0.4 for English datasets), we calculate the F1 score between $A'$ and GT answer $A$ as the final score for $A'$. otherwise, $A'$ gets a zero score. We manually annotate the answer keywords in the GT answer $A$ for \textbf{hotpotwikiqa-mixup}, \textbf{lic-mixup}, \textbf{loogle-CR-mixup}, \textbf{loogle-MR-mixup}, \textbf{loogle-SD-mixup}, \textbf{multifieldqa-en-mixup}, and \textbf{multifieldqa-zh-mixup}. Figure~\ref{fig:ans_keywords_word_blacklst} (a) shows an example, demonstrating how this two-stage calculation helps avoid some inflated high evaluation scores.

When calculating the F1 score between $A'$ and $A$ in the second stage, we exclude common but non-informative words like `the,' `a', `of', and so on. The word blacklist is constructed as follows. We first summarized the word counts in the generations of Llama2-7B-Chat-hf and ChatGLM3-6B-32K on all datasets and chose the top 100 words that matched the GT answer most frequently. Then, we manually annotate the non-informative words from the 100 words to construct the blacklist. Figure~\ref{fig:ans_keywords_word_blacklst} (b) shows an example of how the word blacklist aids in calibrating the evaluation scores.

\section{Evaluation}
\label{sec:evaluation}

% In all of the evaluation experiment, we follow the official implementation of each LLM when inference, avoiding the influence of prompt format and special tokens. Besides, for LLMs whose context window size is smaller than the length of data context, we truncate the data context in the middle, then concatenate the head and the tail of context as input, ensuring the QA instructions are contained completely in the input.

%\subsection{Evaluation Setup}

\paragraph{Models and Inference.} We evaluate 3 commercial and 12 open-source LLMs on \nameshort. Their information is summarized in Table~\ref{tab:LLMs}. %We do not test GPT-4-128k due to the high cost.
We follow the official implementation of all LLMs to conduct their inferences. Greedy sampling is used for generating tokens. For LLMs with a context window size smaller than the length of the data context, we truncate the data context in the middle, and concatenate the head and the tail of the context as input, ensuring that the QA instructions are fully contained within the input.

% Please add the following required packages to your document preamble:
% \usepackage{graphicx}
% \usepackage[table,xcdraw]{xcolor}
% Beamer presentation requires \usepackage{colortbl} instead of \usepackage[table,xcdraw]{xcolor}
\begin{table}[]
\centering
\resizebox{\columnwidth}{!}{%
\begin{tabular}{cccc}
\toprule
{\color[HTML]{333333} Model Name} & SFT & {\color[HTML]{333333} Context Length} & {\color[HTML]{333333} HuggingFace / API Endpoint} \\ \midrule
Llama2-7B-Chat-hf~\cite{touvron2023llama}    & \checkmark & $4k$   & meta-llama/Llama-2-7b-chat-hf \\
Qwen-7B-8k-Chat~\cite{bai2023qwenvl}      & \checkmark & $8k$   & Qwen/Qwen-7B-Chat             \\
Llama3-8B-Instruct~\cite{llama3modelcard}      & \checkmark & $8k$   & meta-llama/Meta-Llama-3-8B-Instruct             \\
Vicuna-7B-16k-v1.5~\cite{zheng2023judging}   & \checkmark & $16k$  & lmsys/vicuna-7b-v1.5-16k      \\
ChatGLM3-6B-32k~\cite{zeng2023glm130b}      & \checkmark & $32k$  & THUDM/chatglm3-6b-32k         \\
BlueLM-7B-32k-Chat~\cite{2023bluelm}   & \checkmark & $32k$  & vivo-ai/BlueLM-7B-Chat-32K    \\
LongChat-7B-32k-v1.5~\cite{togetherAI2023} & \checkmark & $32k$  & lmsys/longchat-7b-v1.5-32k    \\
Qwen2.5-72B-Instruct-128k~\cite{qwen2}            & \checkmark    & $128k$                                   & Qwen/Qwen2.5-72B-Instruct          \\
Meta-Llama-3.1-70B-128k-Instruct~\cite{Llama-3.1-70B-Instruct}            & \checkmark    & $128k$                                   & meta-llama/Llama-3.1-70B-Instruct          \\
Yi-6B-200k~\cite{Yi2023}           &  & $200k$ & 01-ai/Yi-6B-200K              \\
Llama3-8B-1M~\cite{Llama38BGradientInstruct1048k}           & \checkmark & $1048k$ & gradientai/Llama-3-8B-Instruct-Gradient-1048k              \\ \midrule
GPT-4-8k~\cite{openai2023gpt4}             & \checkmark & $8k$   & gpt-4-0613                    \\
GPT-3.5-16k~\cite{ye2023comprehensive}          & \checkmark & $16k$  & gpt-3.5-turbo-1106            \\ 
Moonshot-V1-128k~\cite{moonshot}             & \checkmark & $128k$   & moonshot-V1-128k                    \\ \bottomrule
\end{tabular}%
}
\caption{Information of evaluated LLMs.}
\label{tab:LLMs}
\end{table}

% \begin{figure}[t]
% \centering
% \includegraphics[width=1\linewidth]{Figures/line_perf_levels_fix6.png}
% \caption{\label{fig:avg_perf_length_level}Average scores across all datasets of 12 LLMs at 5 length levels.}
% \end{figure}

% \begin{figure}[t]
% \centering
% \includegraphics[width=1\linewidth]{Figures/satellite_perf_tasks_fix6.png}
% \caption{\label{fig:avg_score_6_subtasks}Average scores across all length levels of 12 LLMs on 5 types of tasks.}
% \end{figure}

% \begin{figure}[t]
% \centering
% \includegraphics[width=1\linewidth]{Figures/bar_perf_sqa_fix6.png}
% \caption{\label{fig:avg_score_sqa}Average scores across all length levels of 12 LLMs on single-hop QA datasets.}
% \end{figure}

% \begin{figure}[t]
% \centering
% \includegraphics[width=1\linewidth]{Figures/bar_perf_mqa_fix6.png}
% \caption{\label{fig:avg_score_mqa}Average scores across all length levels of 12 LLMs on multi-hop QA datasets.}
% \end{figure}

%\paragraph{Inference.} 
%We follow the official implementation of all LLMs to conduct their inferences. Greedy sampling is used for generating tokens. For LLMs with a context window size smaller than the length of the data context, we truncate the data context in the middle, and concatenate the head and the tail of the context as input, ensuring that the QA instructions are fully contained within the input. 

\paragraph{Metrics.}
For all tasks except \textbf{dureader-mixup} and \textbf{cmrc-mixup}, we evaluate the generated answers with our keyword-recall-based F1 metric, utilizing the annotated answer keywords and word blacklist. For \textbf{cmrc-mixup}, we omit the manual annotation of answer keywords since the answers in this dataset is already concise. Therefore, we use the F1 metric with word blacklist. In the case of \textbf{dureader-mixup}, where the GT answer lengths are relatively long, we do not manually annotate the answer keywords and use the ROUGH-L metric with the word blacklist.

% All QA tasks in our benchmark are evaluated on optimized F1 score metric with answer keywords and words blacklist, except for \textbf{dureader-mixup}. Regarding to those datasets with annotated alternative answers, the maximum score among all answers is taken. The ROUGH-L metric with words blacklist is utilized in \textbf{dureader-mixup}, for the reason that the words length of answers is relatively long and the content of most answers are complete paragraphs. In summarization task, standard ROUGH-L metric is utilized.

\subsection{Compare LLMs on \nameshort}

\begin{figure}[t]
%  \vspace{-0.2cm}
\centering
\includegraphics[width=1\linewidth]{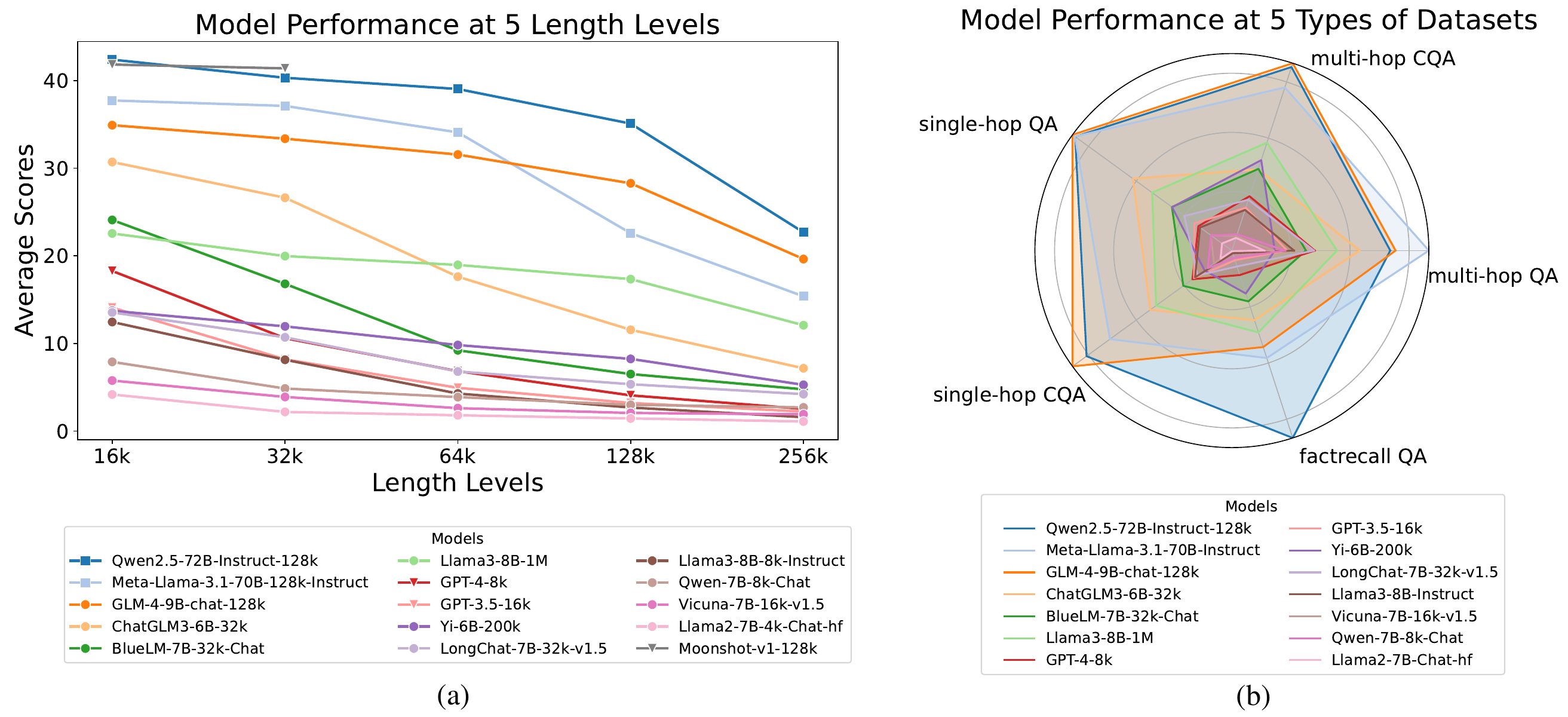}
\vspace{-0.6cm}
\caption{\label{fig:avg_perf_length_level_and_tasks}Overall results on different length levels and types of datasets. (a) Average scores across all datasets of 15 LLMs at 5 length levels. %\marksymbol{*}{black} stands for open-source models of 6-10B; %\marksymbol{square*}{black} stands for large open-source models around 70B;
  The circle markers represent open-source models of 6-9B;
  The square markers represent larger open-source models around 70B;
  The triangle markers represent commercial APIs.
  Note that we only evaluate Moonshot-v1-128k on $16k$ and $32k$ length levels due to the high cost. (b) Average scores across all length levels of 15 LLMs on 5 types of datasets. ``CQA'' refers to QA datasets with CFI.}
\vspace{-0.3cm}
\end{figure}

Figure~\ref{fig:avg_perf_length_level_and_tasks} (a) shows the average scores across all 11 datasets of 15 LLMs at different length levels. We can see that
% (i) Commercial models do not always perform better than open-source models. For instance, ChatGLM3-6B-32k attains the highest accuracy on $16k$ and $32k$.
(i) The commercial Moonshot-v1 and recent large-scale open-source models (Qwen2.5-72B, Llama-3.1-70B) achieve the best performances. Notably, both Qwen2.5-72B and Moonshot-v1 obtain average scores exceeding 40 at $16k$ and $32k$ lengths. (ii) Among open-source models with parameter sizes in the 6-9B range, GLM-4-9B achieves the best performance, even outperforming Llama-3.1-70B on longer lengths ($128k$, $256k$).
(iii) Models exhibit distinct score trends, resulting in different relative rankings across different length levels. 
For example, the model with the largest context window size, Llama3-8B-1M, exhibits one of the slowest decline of performance from $16k$ to $128k$. Specifically, its scores at the length level $16k$ is lower than ChatGLM3-6B-32k and BlueLM-7B-32k-Chat. Nevertheless, as the length of input context increases, Llama3-8B-1M retains a higher score than these two models that need to truncate the input context. The similar phenomenon can be observed between Yi-6B-200k and two GPTs, as well as between GLM-4-9B and Llama-3.1-70B.
(iv) Models exhibit a sharp performance drop, often occurring prior to or when the context length exceeds their supported context length. Some models experience a sharp decline in performance once the context length exceeds their supported context length. For instance, Qwen2.5-72B-128k, GLM-4-9B-128k, and ChatGLM3-6B-32k exhibit a sharp performance drop after $128k$, $128k$, and $32k$, respectively. In contrast, some other models encounter this sharp performance drop earlier. For instance, Llama-3.1-70B-128k exhibit a sharp performance drop after $64k$, despite its stated support for $128k$ context.

Figure~\ref{fig:avg_perf_length_level_and_tasks} (b) shows the average scores across all 5 length levels on 5 task types. We can see that (i) LLMs attain lower scores on multi-hop QA tasks compared to single-hop QA tasks. (ii) Confusing facts insertion adds complexity to the tasks, particularly evident in single-hop QA and single-hop confusion QA.
See Appendix~\ref{appendix:B} for more detailed results and Appendix~\ref{sec:error_cases} for example failure cases.

\subsection{Ablation Study of \nameshort Techniques}

\paragraph{Confusing facts insertion.}
% 总表+大海捞针图

Table~\ref{tab:res_ablt_study_ci_kpr_hot}, \ref{tab:res_ablt_study_ci_kpr_mqae}, and \ref{tab:res_ablt_study_ci_kpr_mqaz} show the scores of multiple LLMs on dataset with and without CFI. We can see that (i) On \textbf{multifieldqa-en-mixup} and \textbf{multifieldqa-zh-mixup}, CFI leads to a notable degradation in the scores of LLMs. However, CFI in the \textbf{hotpotwikiqa-mixup} dataset does not result in severe degradation. (ii)  Table~\ref{tab:res_ablt_study_ci_kpr_mqae} and \ref{tab:res_ablt_study_ci_kpr_mqaz} show that a strong model, ChatGLM3-6B-32k, exhibits the most substantial score degradation on data with CFI. For instance, the score of ChatGLM3-6B-32k degrades from $41.46$ to $31.97$ (a degradation of $9.49$) on the $16k$ length level of \textbf{multifieldqa-en-mixup}, while the score degradation of other 5 LLMs falls within the range $\left[0.47, 4.89\right]$. This observation suggests that current powerful LLMs may even be more susceptible to confusing information in the context. Future research is needed to enhance the models' ability to discern information that appears similar but is in fact unrelated. (iii) 
As the length of the input context increases, the score degradation becomes smaller. This phenomenon can be attributed to two factors: the truncation of confusing facts and a decrease in baseline performance.

\paragraph{Keyword and phrase replacement.}
% 总表

The technique of KPR aims to eliminate the knowledge leakage and common-sense memorization of LLMs. Intuitively, for datasets sourced from Wikipedia and other widely used corpus, the risk of knowledge leakage is higher.
From the results in Table~\ref{tab:res_ablt_study_ci_kpr_hot}, \ref{tab:res_ablt_study_ci_kpr_mqae}, and \ref{tab:res_ablt_study_ci_kpr_mqaz}, we observe that: (i) KPR brings notable degradation of LLM scores on these three datasets suggesting that knowledge leakage exists in open-source corpus and can be mitigated by KPR. (ii) The extent of degradation is relatively consistent across different length levels.

We illustrate the knowledge leakage issue and the impact of KPR in Table~\ref{tab:comp_res_wwo_ctx}. Specifically, we compare three settings: (i) Directly querying the LLMs to answer the question without the context (``direct (w.o. KPR)''). (ii) Applying KPR to the QA pair, and directly querying the LLMs without the context (``direct (w. KPR)''). (iii) Applying KPR to the QA pair and the context, and querying the LLMs to answer the question with the context (``w. context (w. KPR)'').

Table~\ref{tab:comp_res_wwo_ctx} shows that without KPR, some LLMs can achieve a considerable score even without context. For instance, Yi-6B-200k and ChatGLM3-6B-32k achieve scores of 16.11 and 12.24, respectively, through memorization or common-sense knowledge. Applying KPR decreases the score without context (6.06 for Yi-6B-200k and 4.96 for ChatGLM3-6B-32k). This helps mitigate the influence of memorization or common-sense knowledge on the assessment of long-context understanding ability.

% Please add the following required packages to your document preamble:
% \usepackage{multirow}
% \usepackage{graphicx}
\begin{table}[t]
    \vspace{-0.2cm}
\centering
%\resizebox{\columnwidth}{!}{%
\begin{tabular}{c|c|ccccc}
\toprule
\multirow{2}{*}{Model Name} &
  \multirow{2}{*}{Ablation} &
  \multicolumn{5}{c}{\textbf{hotpotwikiqa-mixup}} \\ \cmidrule{3-7} 
 &              & \multicolumn{1}{c}{$16k$} & \multicolumn{1}{c}{$32k$} & \multicolumn{1}{c}{$64k$} & \multicolumn{1}{c}{$128k$} & \multicolumn{1}{l}{$256k$} \\ \midrule
\multirow{3}{*}{Llama2-7B-Chat-hf} &
  direct (w. KPR) &
  \multicolumn{5}{c}{2.43} \\ 
 &
  direct (w.o. KPR) &
  \multicolumn{5}{c}{3.52} \\ 
 &
  w. context (w. KPR) &
  \multicolumn{1}{c}{3.99} & 				
  \multicolumn{1}{c}{1.30} &
  \multicolumn{1}{c}{1.84} &
  \multicolumn{1}{c}{0.81} &
  0.75 \\ \midrule
\multirow{3}{*}{ChatGLM3-6B-32k} &
  direct (w. KPR) &
  \multicolumn{5}{c}{4.96} \\ 
 &
  direct (w.o. KPR) &
  \multicolumn{5}{c}{12.24} \\  				
 & w. context (w. KPR) & \multicolumn{1}{c}{16.98} & \multicolumn{1}{c}{14.76} & \multicolumn{1}{c}{9.02} & \multicolumn{1}{c}{8.31}   & 6.68                       \\ \midrule
% \multirow{3}{*}{LongChat-7B-32k-v1.5} &
%   direct (w. KPR) &
%   \multicolumn{5}{c}{3.84} \\ 
%  &
%   direct (w.o. KPR) &
%   \multicolumn{5}{c}{8.20} \\ 
%  &
%   w. context (w. KPR) &
%   \multicolumn{1}{c}{11.57} & 			
%   \multicolumn{1}{c}{10.71} &
%   \multicolumn{1}{c}{4.77} &
%   \multicolumn{1}{c}{5.49} &
%   2.37 \\ \midrule
% \multirow{3}{*}{Llama2-7B-32k-Instruct} &
%   direct (w. KPR) &
%   \multicolumn{5}{c}{0.76} \\ 
%  &
%   direct (w.o. KPR) &
%   \multicolumn{5}{c}{2.48} \\ 
%  &
%   w. context (w. KPR) &
%   \multicolumn{1}{c}{3.54} & 		
%   \multicolumn{1}{c}{2.31} &
%   \multicolumn{1}{c}{2.20} &
%   \multicolumn{1}{c}{1.86} &
%   1.62 \\ \midrule
\multirow{3}{*}{Yi-6B-200k} &
  direct (w. KPR) &
  \multicolumn{5}{c}{6.06} \\ 
 &
  direct (w.o. KPR) &
  \multicolumn{5}{c}{16.11} \\  				
 & w. context (w. KPR) & \multicolumn{1}{c}{23.55} & \multicolumn{1}{c}{18.94} & \multicolumn{1}{c}{9.94}  & \multicolumn{1}{c}{7.66}   & 2.01                       \\ \bottomrule
\end{tabular}%
%}
\caption{Ablation results for KPR. ``direct (w. KPR)'': Apply KPR and direct query without context; ``direct (w.o. KPR)'': Direct query without context; ``w. context (w. KPR)'': Apply KPR and query with context. Note that there is only one result in the first two rows in each section of the table, since the results of direct querying without context do NOT depend on the context length.}
\label{tab:comp_res_wwo_ctx}
\vspace{-0.3cm}
\end{table}

\paragraph{Case study on the fact-recall tasks.}

The \textbf{factrecall-en} and \textbf{factrecall-zh} datasets are constructed to evaluate the enhanced NIAH~\citep{needle2023} ability. The traditional NIAH evaluation is basically a retrieval task, asking LLMs to find the answer or passkey in long context, which is too simple for majority of LLMs that they can easily get high scores after task oriented training. Therefore we enhance the NIAH evaluation with CFI and KPR to assess LLM's positional consistency of retrieval while challenging their comprehension and anti-interference abilility.

\begin{figure}[ht]
%\vspace{-0.1cm}
\centering
\includegraphics[width=1\linewidth]{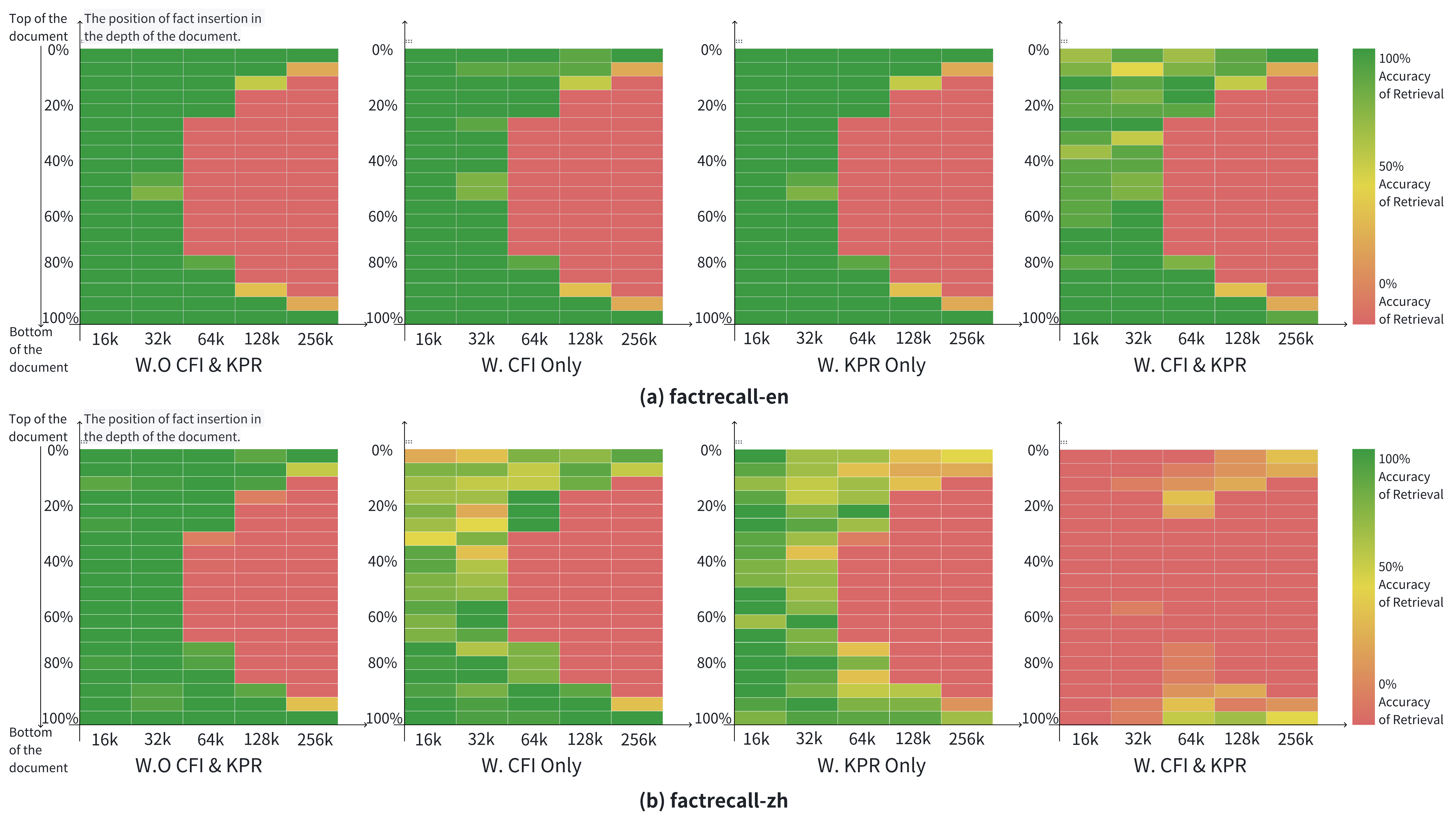}
\vspace{-0.7cm}
\caption{\label{fig:needle_haystack}Ablation results of the ``needle in a haystack'' task on ChatGLM3-6B-32k. (a) \textbf{factrecall-en}. (b) \textbf{factrecall-zh}. In each of (a)(b), from left to right, the four sub-figures show the results of w.o. ``CFI and KPR'', ``w. CFI only'', ``w. KPR only'', and ``w. both CFI and KPR'', respectively. These results illustrate that CFI and KPR are effective in improving the task difficulty.}
\end{figure}

We show the ablation results of CFI and KPR in Figure~\ref{fig:needle_haystack} and Table~\ref{tab:res_factrecall_enzh}. From the first column in Figure~\ref{fig:needle_haystack}, we can see that ChatGLM3-6B-32k attains high accuracy on datasets without CFI and KPR, as long as the input context length is within its context size (32k). However, when either CFI (second column of sub-figure) or KPR (third column sub-figure) is applied, the retrieval accuracy decreases. The accuracy experiences a more severe degradation when both CFI and KPR are applied, particularly evident in \textbf{factrecall-zh}, where a performance collapse is observed (See Appendix~\ref{sec:error_cases} for an example failure case). This indicates that there is room for improvement in the model's ability to accurately identify a specific piece of information from a long context in the presence of interference.
%and the original ``needle in a haystack'' could not be suit for reasonably evaluating long context capability.

\paragraph{Keyword-recall-based metric.}
For a given length level $L_d$ of the dataset, if the single key information is uniformly distributed in the context, an LLM with a context window size $L_m$ can only observe the key information for approximately $\frac{L_m}{L_d}$ of the time. 
Thanks to our KPR technique, we can reasonably expect that the LLM cannot get the correct answer through memorization or common-sense knowledge. Furthermore, in our free-form QA tasks, unlike in multiple-choice settings, the LLM cannot easily guess the correct answer. Therefore, we would not expect to see a metric score much higher than $\frac{L_m}{L_d}$. In other words, $\frac{L_m}{L_d}$ is the ideal score assuming that all answers are correctly extracted from and only from the context, and the metric accurately measures the answer correctness.

However, as shown in Table~\ref{tab:F1_metric_scores}, when using the original F1 metric, the metric score can be significantly higher than $\frac{L_m}{L_d}$ due to the undesired matching of non-keywords and non-informative words. For example, ChatGLM3-6B-32k achieves a score of $26.43\%$ on the $256k$ length level of the \textbf{cmrc-mixup} dataset, which greatly exceeds $\frac{L_m}{L_d}=12.5\%$. In contrast, our keyword-recall-based metric with the word blacklist is a more meaningful metric, as its scores are more aligned with our expectations (i.e., smaller than $\frac{L_m}{L_d}$).

\begin{table}[ht]
\centering
%\resizebox{\linewidth}{!}{
\begin{tabular}{cccccc}
\toprule
Metric    & $16k$ & $32k$ & $64k$ & $128k$ & $256k$ \\\midrule
reference $\frac{L_m}{L_d}$   & 100           & 100       & 50.00          & 25.00          & 12.50       \\
original 		       & 66.49		    & 59.99       & 38.71          & 31.76          & 26.43       \\
w. answer keywords       & 57.67		   & 52.18       & 28.92         & 21.07          & 15.45       \\
w. answer keywords + word blacklist       & 51.21	    & 46.34       & 20.71          & 14.16          & 8.38       \\ \cmidrule(l){1-6}
\end{tabular}
%}
\caption{Metric scores of ChatGLM3-6B-32k on \textbf{cmrc-mixup}. The score inflation is suppressed with keyword-recall-based metric design.}
\label{tab:F1_metric_scores}
\end{table}

% \section{Conclusion}

\section{Limitations}
\label{others}

\nameshort does not encompass other task types such as summarization. Additionally, due to the high cost, we do not test some of the most recent LLMs, such as GPT-4-128k, GPT-4o, and so on. As we release all the test data, one can intentionally overfit the benchmark by training on the test data to get a high score. In this case, training on LV-Eval datasets with KPR might lead to mistakes in common-sense knowledge, resulting in a very unreliable evaluation. 

A worth-noting issue about KPR is that as KPR modifies the fact, there exist some cases where the model identifies factual errors and then insists on providing a common-sense response. We view these cases as an issue with the model's instruction-following ability, as our prompt explicitly states that the answer should be solely based on the context instead of its existing knowledge.

Our CFI technique relies on manual revision of the confusing facts to ensure the benchmark quality. A possible way to alleviate the human efforts when applying the CFI technique to future corpus might be leveraging stronger LLMs to substitute manual revision efforts.

%Furthermore, a full evaluation on \nameshort can cause a large token overhead (about 700M tokens for GPT-4's tokenizer), leading to considerable carbon emissions.

\section*{Acknowledgments}
We thank the insightful discussion with Kai Chen and Songyang Zhang from Shanghai Artificial Intelligence Laboratory. Thanks to all of the annotators who contribute to the project. We thank Hanling Zhang for helping run the experiment for Moonshot. We thank Qin Cheng from Yale University for presenting our poster at the Conference on Language Modeling (COLM 2025).

\bibliography{iclr2025_conference}
\bibliographystyle{iclr2025_conference}

\clearpage
\appendix

%\section*{\centering\LARGE Appendix}

\part{Appendix}
\parttoc
\clearpage

\renewcommand{\thefigure}{A\arabic{figure}}
\renewcommand{\thetable}{A\arabic{table}}
\renewcommand{\theequation}{A\arabic{equation}}

\begin{figure}[h]
\centering
\includegraphics[width=0.84\linewidth]{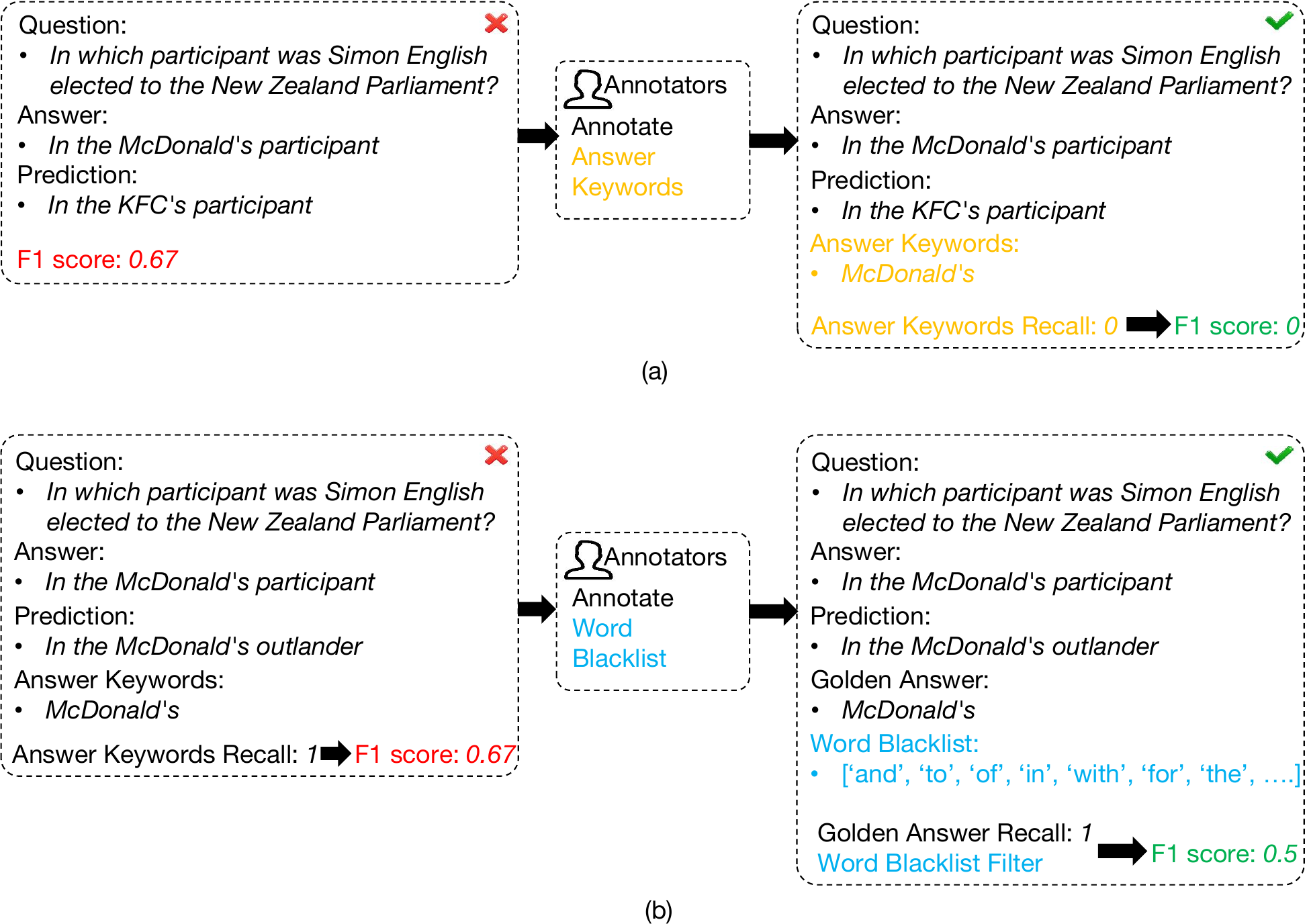}
\caption{\label{fig:ans_keywords_word_blacklst}Keyword-recall-based two-stage metric calculation. (a) The vanilla F1 score (red) is inflated high. With the keyword recall-based metric, the final score is set to zero due to the low recall of answer keywords. (b) The vanilla F1 score (red) is inflated high due to irrelevant words. By filtering of blacklisted words, the final score is better calibrated.}
\end{figure}

\begin{figure}[h]
\centering
\includegraphics[width=0.84\linewidth]{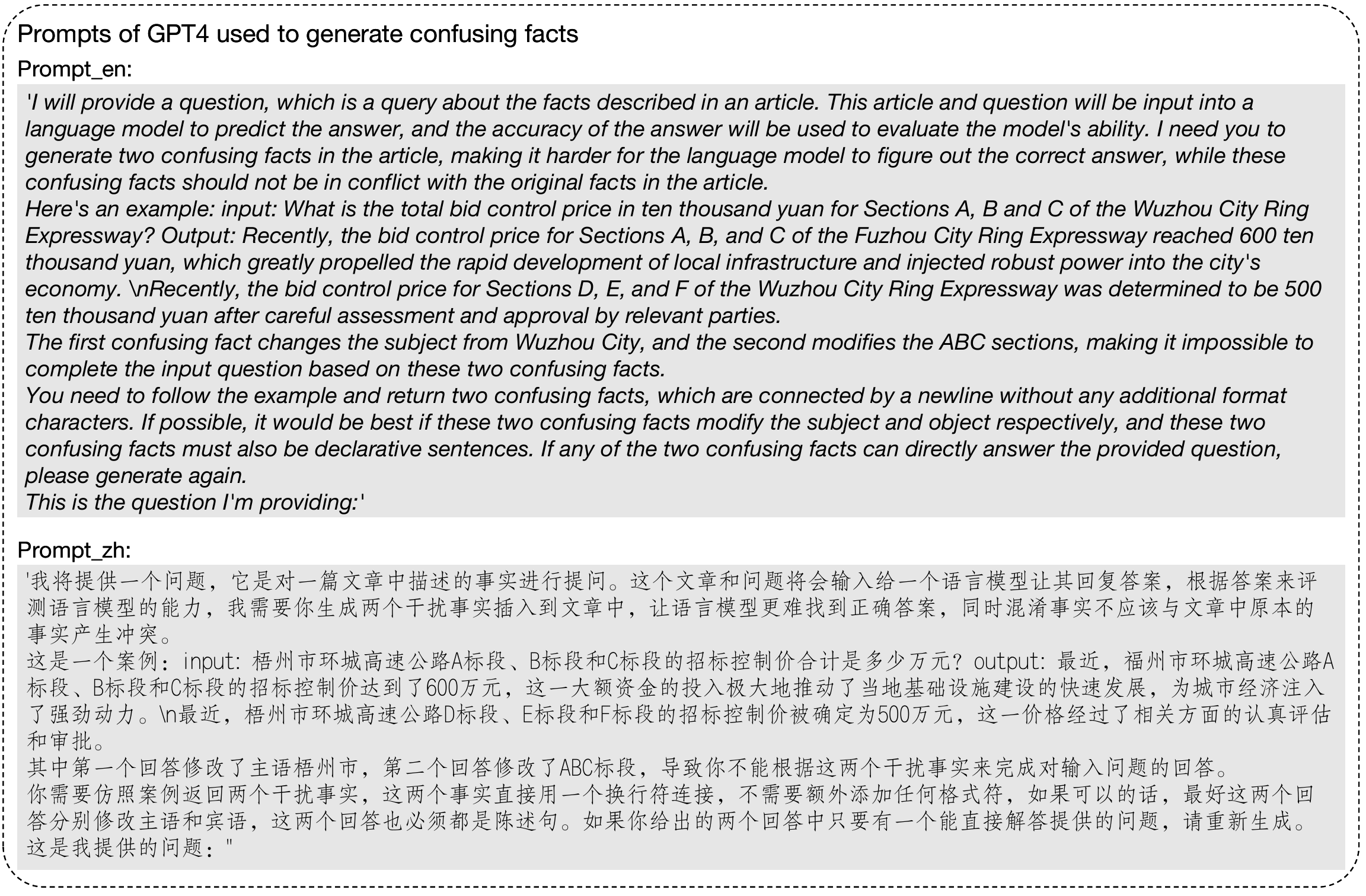}
\caption{\label{fig:confusing_fact_prompt}Prompts used for GPT-4 to generate confusing facts.}
\end{figure}

\section{Detailed Construction of QA Pairs}
\label{appendix:A}

\paragraph{Multi-hop QA.}
In a multi-hop QA task, the reasoning to derive the answer needs to gather multiple pieces of information from various locations in the context. We construct four multi-hop QA datasets: \textbf{dureader-mixup}, \textbf{loogle-CR-mixup}, \textbf{loogle-MR-mixup}, and \textbf{hotpotwikiqa-mixup}. 
\begin{itemize}

\item \textbf{hotpotwikiqa-mixup} is originated from two Wikipedia-based multi-hop QA datasets: HotpotQA and 2WikiMultihopQA. HotpotQA contains 112,779 2-hop questions that are written by native speakers according to two given paragraphs as the context. 2WikiMultihopQA contains 192,606 5-hop questions that are synthesized using manually designed templates to prevent shortcut solutions. We select 124 samples from the two datasets.
  
\item \textbf{loogle-MR-mixup} and \textbf{loogle-CR-mixup} originate from LooGLE's Long-dependency QA task, specifically the \textit{Multiple information Retrieval} and \textit{Comprehension and Reasoning} subtasks. The \textit{Multiple information Retrieval} task requires aggregation of the evidence that can be directly located in original sentences, while the \textit{Comprehension and Reasoning} task contains implicit evidence within the context, it requires multi-step reasoning to get the correct answers. We select  139 and 99 questions for \textbf{loogle-MR-mixup} and \textbf{loogle-CR-mixup}, respectively.

\item \textbf{dureader-mixup} is built from the DuReader dataset. We first randomly select 200 instances and then manually remove 24 samples whose answers are longer than 360 words.
\end{itemize}

\paragraph{Single-hop QA.}
In a single-hop QA task, only a single evidence in the context is needed to derive the answer. We construct seven single-hop QA datasets: \textbf{lic-mixup}, \textbf{loogle-SD-mixup}, \textbf{cmrc-mixup}, \textbf{multifieldqa-en-mixup}, \textbf{multifieldqa-zh-mixup}, \textbf{factrecall-en}, and \textbf{factrecall-zh}.

\begin{itemize}
\item \textbf{lic-mixup} is originated from the \verb+Long-instruction-en2zh+ dataset on Hugging Face. \verb+Long-instruction-en2zh+ contains 8,000+ high-quality Chinese multi-doc QA data translated from English. We selected 197 QA pairs and their corresponding documents as supporting data, while the remaining documents serve as distracting data for context mixing.
\item \textbf{loogle-SD-mixup} contains 160 unique QA pairs and 800 documents originated from the short-dependency QA task in LooGLE.
\item \textbf{cmrc-mixup} is derived from the CMRC 2018 Public Datasets, designed for Chinese machine reading comprehension. 
It contains $\sim$$20k$ questions annotated on Wikipedia documents by human experts.
We manually pick 200 QA pairs and their corresponding documents as supporting QA pairs and documents. 
\item \textbf{multifieldqa-en-mixup} and \textbf{multifieldqa-zh-mixup} are built from the MultiFieldQA datasets in Long-Bench.
We manually remove questions that can be answered using common-sense knowledge without referring to the context, and eventually get 101 and 133 unique QA pairs for  \textbf{multifieldqa-en-mixup} and \textbf{multifieldqa-zh-mixup}, respectively.
\item \textbf{factrecall-en} and \textbf{factrecall-zh} are two synthetic datasets designed to assess the LLMs' ability to identify a small piece of evidence (``fact'') located at various locations within a lengthy context.
As shown in Figure~\ref{fig:sample_factrecall_en}~\ref{fig:sample_factrecall_zh}, we write one English fact-question-answer pair for \textbf{factrecall-en} and one Chinese fact-question-answer pair for \textbf{factrecall-zh}. 
distracting documents are sourced from  \textit{PG-19} dataset (English) and the book of \textit{Journey to the West} (Chinese) to create five contexts of different length levels.
For each context, we generate 200 documents by inserting the fact at 200 evenly spaced positions within the context.
\end{itemize}

\section{Annotation Details}
\label{sec:annotation}
\begin{itemize}
\item \textbf{Annotators}: We hire a total of five annotators, including three master students involved in LLM research and two master students majored in linguistics. They are hired to work on-site as full-time annotators.

\item \textbf{Annotation time}: On average, annotators worked 8 hours per day. (1) For CFI, six datasets comprising a total of 557 instances of confusing facts were reviewed, with the verification process completed by two annotators over a period of three days. This task was assigned to master's students in linguistics to ensure semantic consistency. (2) For KPR, six datasets containing 1,924 pairs were processed. Replacing key words and phrases in the Chinese datasets involved five individuals over three days, whereas the English datasets required two individuals over two days. (3) For answer keyword annotation, nine datasets comprising 955 instances of answer keywords were annotated by two individuals within a single day.

\item \textbf{Annotation guidelines}: We provided guidelines and examples to the five annotators and conducted a trial annotation in accordance with these guidelines to ensure they understand the guidelines. The guidelines are summarized as follows. For CFI, two types of cases require manual modification: (a) The confusing facts generated by GPT-4 fail to produce interference; (b) The confusing facts generated by GPT-4 conflict with the original facts. 

For KPR, the guidelines are: (a) Ensure that the replaced words or phrases are distinct from the original and not synonyms; (b) Maximize the differences between the revised and original sentences by replacing as many words as possible; (c) Prioritize the replacement of words that do not have synonyms, as it is challenging to ensure that all synonyms are correctly replaced for words with multiple synonyms; (d) After replacement, the resulting statement may be inconsistent with common knowledge; (d) After replacement, it's OK that the resulting statement may be inconsistent with common knowledge, as long as the answer can be derived from the context and that all information related to this answer is consistent within the context.
\end{itemize}

\section{Detailed Evaluation Results}
\label{appendix:B}
%The information of all LLMs are listed in Table~\ref{tab:LLMs}.

\begin{figure}[ht]
\centering
\includegraphics[width=0.84\linewidth]{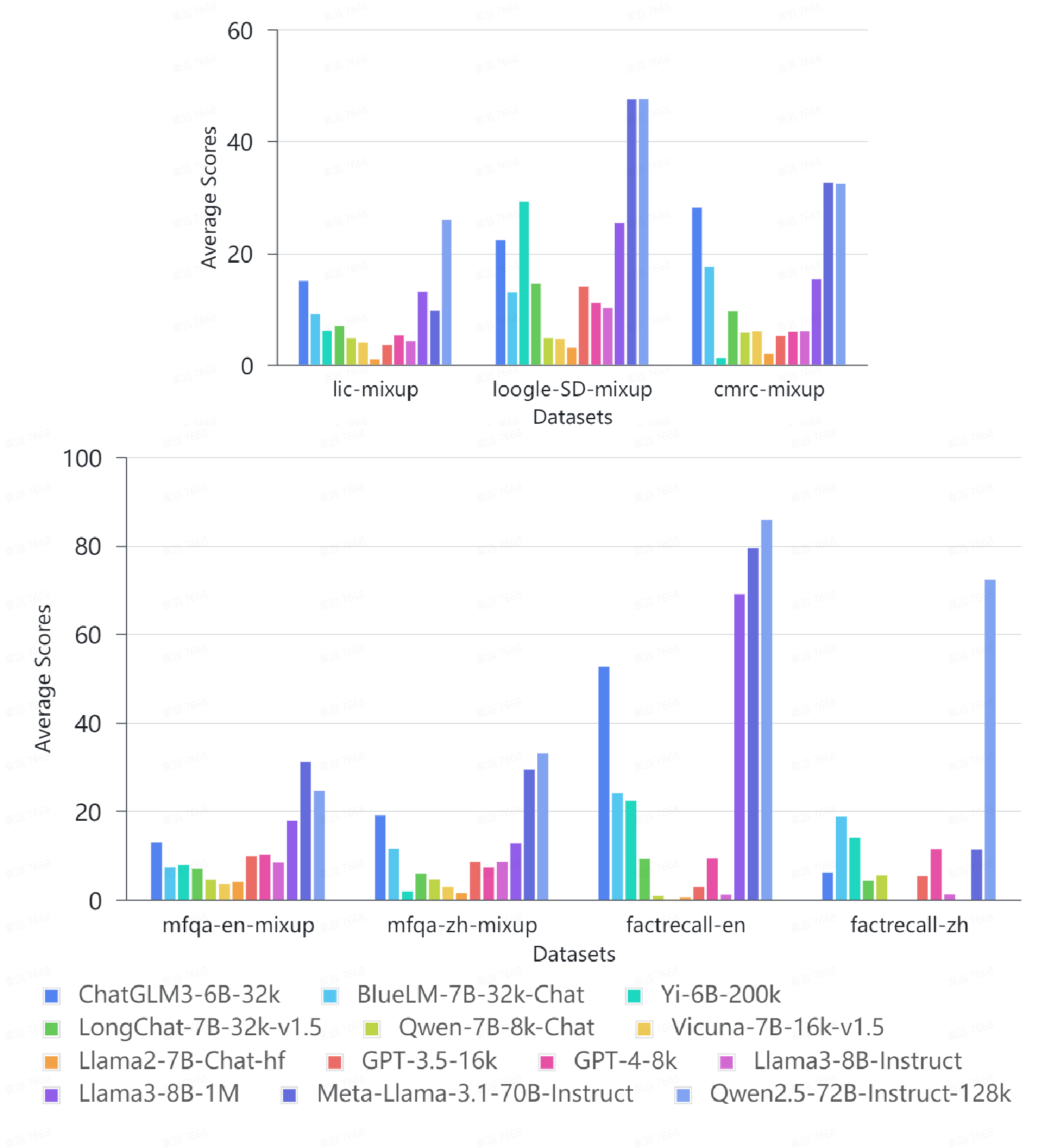}
\caption{\label{fig:avg_score_sqa}Average scores across all length levels of 15 LLMs on single-hop QA datasets.}
\end{figure}

\begin{figure}[h]
\centering
\includegraphics[width=0.84\linewidth]{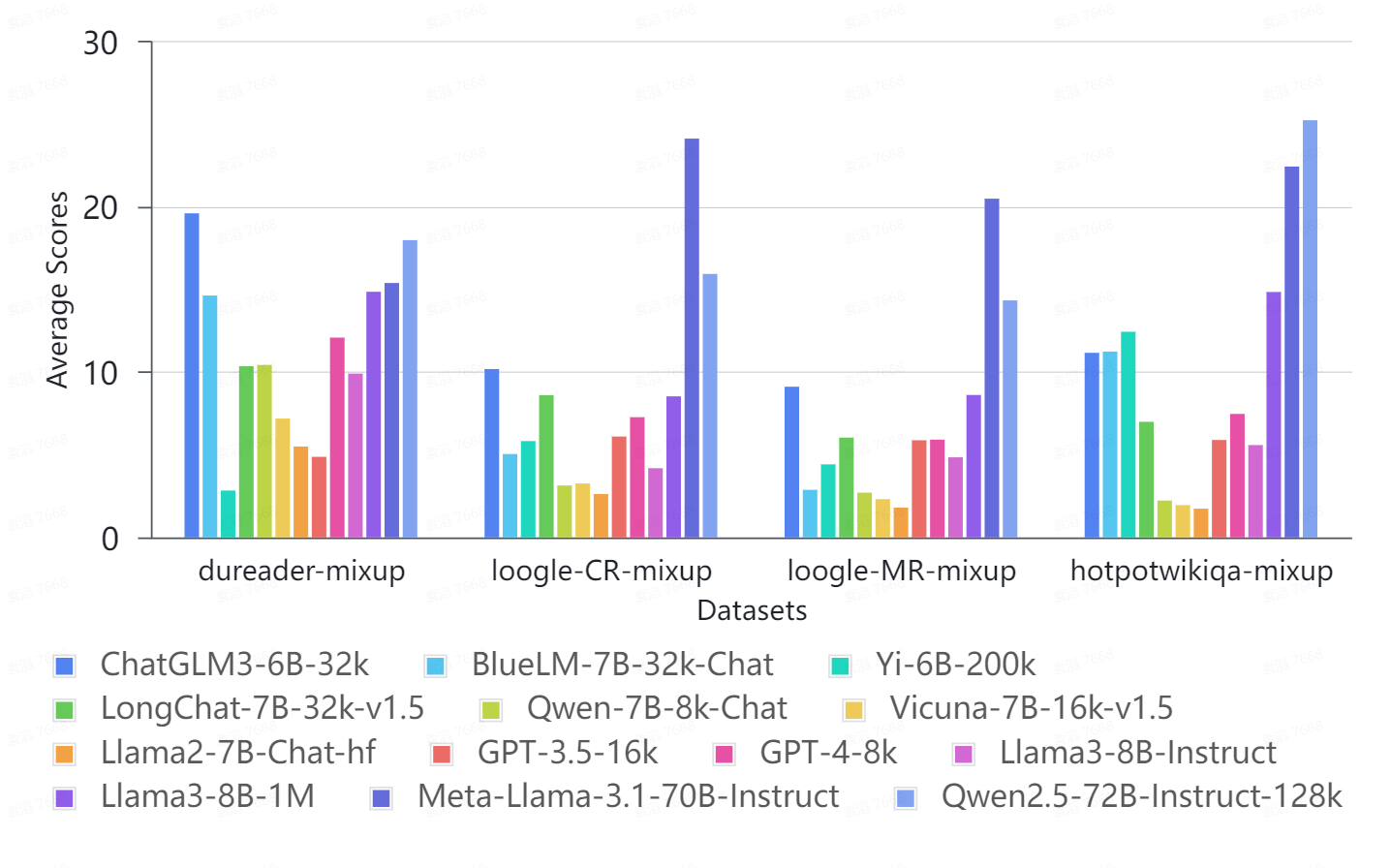}
\caption{\label{fig:avg_score_mqa}Average scores across all length levels of 15 LLMs on multi-hop QA datasets.}
\end{figure}

The detailed results on each dataset of the single-hop QA task type and multi-hop QA task type are shown in Figure~\ref{fig:avg_score_sqa} and Figure~\ref{fig:avg_score_mqa}, respectively. We can see that (i) Among the multi-hop QA datasets, \textbf{loogle-CR-mixup} and \textbf{loogle-MR-mixup} are particularly challenging. Future research is needed to improve the ability to aggregate multiple pieces of evidence from a long context with distracting and confusing facts. (ii) For single-hop QA datasets, as expected, LLMs can achieve higher scores on datasets without CFI, including \textbf{loogle-SD-mixup} and \textbf{cmrc-mixup}. (iii) Several LLMs, namely ChatGLM3-6B-32k, BlueLM-7B-32k-Chat, and Yi-6B-200k, can achieve relatively high scores on \textbf{factrecall-en}. This indicates that the NIAH task might not be challenging enough, emphasizing the need to evaluate LLMs on other tasks, particularly multi-hop QA datasets.
(iv) The performance gap between LLMs on \textbf{factrecall-en} and \textbf{factrecall-zh} is especially large, and some open-source LLMs with relatively small context sizes, namely Llama2-7B-Chat-hf (4k context window size), Qwen-7B-8k-Chat, and Vicuna-7B-16k-v1.5, even get near-zero scores.
(v) A few LLMs have unbalanced performances on Chinese and English datasets, as illustrated by the results on \textbf{multifieldqa-en-mixup} and \textbf{multifieldqa-zh-mixup}. 
The detailed scores of all models on 5 length levels of all sub-datasets are shown in Table~\ref{tab:overall_res_mhqa}~\ref{tab:overall_res_shqa}.

\begin{table}[ht]
\centering
\resizebox{\columnwidth}{!}{%
\begin{tabular}{cccccccccccccccc}
\toprule
Dataset &
  \multicolumn{1}{c}{Len.} &
  \multicolumn{1}{c}{Ch.} &
  \multicolumn{1}{c}{Bl.} &
  \multicolumn{1}{c}{Yi.} &
  \multicolumn{1}{c}{Lo.} &
  \multicolumn{1}{c}{Qw.} &
  \multicolumn{1}{c}{Vi.} &
  \multicolumn{1}{c}{Ll.4} &
  \multicolumn{1}{c}{Ll.8} &
  \multicolumn{1}{c}{Ll.M} &
  GPT3. &
  GPT4. &
  \multicolumn{1}{c}{Ll.31} &
  \multicolumn{1}{c}{Qw.25} &
  \multicolumn{1}{c}{Ms.} \\ \midrule
\multicolumn{1}{c|}{\multirow{5}{*}{\begin{tabular}{c}\textbf{dureader}\\\textbf{-mixup}\end{tabular}}}     & \multicolumn{1}{c|}{$16k$}  & 23.99 & 19.40  & 2.87  & 13.44 & 11.82     & 9.67  & 7.21 & 16.39 & 18.06 & 8.01  & 19.14 & 20.22 & 18.28 & 21.46\\
\multicolumn{1}{c|}{}                                     & \multicolumn{1}{c|}{$32k$}  & 25.21 & 19.74 & 2.98  & 11.57  & 12.80  & 7.65  & 5.42 & 13.08 & 15.86 & 5.26  & 13.64 & 22.38 & 19.16& 21.00\\
\multicolumn{1}{c|}{}                                     & \multicolumn{1}{c|}{$64k$}  & 22.01 & 14.44 & 2.88  & 9.23   & 10.48 & 6.62  & 5.59 & 10.24 & 15.16 & 4.26  & 12.66 & 19.22& 17.82& -\\
\multicolumn{1}{c|}{}                                     & \multicolumn{1}{c|}{$128k$} & 17.94 & 10.95 & 2.36  & 9.51   & 8.15  & 6.25  & 4.78 & 5.30 & 14.46 & 3.30  & 8.19  & 9.90& 18.24& -\\
\multicolumn{1}{c|}{}                                     & \multicolumn{1}{c|}{$256k$} & 8.72  & 8.51  & 3.06  & 7.96   & 8.65  & 5.70   & 4.45 & 4.46 & 10.64 & 3.50 & 6.71  & 5.11& 16.27& -\\ \midrule
\multicolumn{1}{c|}{\multirow{5}{*}{\begin{tabular}{c}\textbf{loogle-CR}\\\textbf{-mixup}\end{tabular}}}   & \multicolumn{1}{c|}{$16k$}  & 14.41 & 9.01  & 8.25  & 11.25  & 5.48  & 5.00     & 3.69 & 8.63&12.56 & 10.04 & 12.68 & 30.05 & 20.33 & 25.02\\
\multicolumn{1}{c|}{}                                     & \multicolumn{1}{c|}{$32k$}  & 14.10  & 7.36  & 8.83  & 11.17  & 3.30   & 4.25  & 3.29 &8.74&11.05 & 8.39  & 10.40 & 26.24 & 20.25 & 23.29  \\
\multicolumn{1}{c|}{}                                     & \multicolumn{1}{c|}{$64k$}  & 9.92  & 3.81  & 4.73  & 9.31   & 3.82  & 3.76  & 3.13 &2.78&8.64 & 5.58  & 6.48  & 26.81 & 19.08 & - \\
\multicolumn{1}{c|}{}                                     & \multicolumn{1}{c|}{$128k$} & 6.95  & 2.40   & 4.05  & 6.19   & 1.14  & 1.99  & 2.19 &0.26&5.81 & 3.08  & 2.83  & 20.72 & 12.56 & - \\
\multicolumn{1}{c|}{}                                     & \multicolumn{1}{c|}{$256k$} & 5.46  & 2.60   & 3.23  & 5.03   & 1.94  & 1.28  & 0.81 &0.49&4.54 & 3.37  & 3.91  & 16.61 & 7.31 & - \\ \midrule
\multicolumn{1}{c|}{\multirow{5}{*}{\begin{tabular}{c}\textbf{loogle-MR}\\\textbf{-mixup}\end{tabular}}}   & \multicolumn{1}{c|}{$16k$}  & 15.83 & 4.90   & 6.94  & 10.53  & 4.93  & 5.17  & 3.37 &10.39&13.73 & 12.95 & 12.24  & 23.60 & 19.14 & 20.62\\
\multicolumn{1}{c|}{}                                     & \multicolumn{1}{c|}{$32k$}  & 11.62 & 3.14  & 7.67  & 9.51   & 2.95  & 3.83  & 2.20 &7.14&10.9  & 7.03  & 7.83  & 22.54 & 16.12 & 19.63 \\
\multicolumn{1}{c|}{}                                     & \multicolumn{1}{c|}{$64k$}  & 7.00     & 1.68  & 2.69  & 3.04    & 2.37  & 0.96  & 2.05 &3.89
&7.82 & 6.23  & 6.26  & 21.57 & 16.52 & - \\
\multicolumn{1}{c|}{}                                     & \multicolumn{1}{c|}{$128k$} & 7.24  & 2.46  & 3.44  & 4.05    & 1.80   & 0.55  & 1.04 &2.37&5.93 & 2.13  & 2.30   & 18.75 & 12.06 & - \\
\multicolumn{1}{c|}{}                                     & \multicolumn{1}{c|}{$256k$} & 3.82  & 2.19  & 1.32  & 3.01    & 1.46  & 1.06  & 0.33 &0.4&4.63 & 1.00     & 0.90   & 15.83 & 7.74 & - \\ \midrule
\multicolumn{1}{c|}{\multirow{5}{*}{\begin{tabular}{c}\textbf{hotpotwikiqa}\\\textbf{-mixup}\end{tabular}}} & \multicolumn{1}{c|}{$16k$}  & 16.98 & 19.31 & 23.55 & 11.57    & 2.78  & 2.63  & 3.99 &12.14&17.67 & 11.96 & 13.51  & 27.53 & 31.69 & 30.08\\
\multicolumn{1}{c|}{}                                     & \multicolumn{1}{c|}{$32k$}  & 14.76 & 14.07 & 18.94  & 10.71    & 1.89  & 2.19  & 1.30 &7.37&17.17 & 6.66  & 10.62  & 30.86 & 31.96 & 28.93\\
\multicolumn{1}{c|}{}                                     & \multicolumn{1}{c|}{$64k$}  & 9.02 & 9.63 & 9.94  & 4.77    & 2.27  & 2.05  & 1.84 &2.34&13.37 & 3.27  & 6.67  & 27.57 & 26.80 & - \\
\multicolumn{1}{c|}{}                                     & \multicolumn{1}{c|}{$128k$} & 8.31   & 7.71  & 7.66  & 5.49   & 2.37  & 1.04  & 0.81 &3.86&15.02 & 4.23  & 4.13  & 17.97 & 21.46 & - \\
\multicolumn{1}{c|}{}                                     & \multicolumn{1}{c|}{$256k$} & 6.68  & 5.40   & 2.01  & 2.37   & 1.82  & 1.85  & 0.75 &2.17&10.88 & 3.30  & 2.36  & 8.07 & 14.07 & - \\ \bottomrule

% \multicolumn{1}{c|}{\multirow{5}{*}{summary-en}}         & \multicolumn{1}{c|}{$16k$}  & 30.69 & 20.51 & 23.66 & 23.53 & 25.19 & 19.04 & 20.51 & 22.44 & 22.38 & 25.18 \\
% \multicolumn{1}{c|}{}                                     & \multicolumn{1}{c|}{$32k$}  & 28.64 & 17.79 & 25.59 & 22.14 & 26.18 & 17.63 & 20.12 & 20.55 & 21.75 & 24.15 \\
% \multicolumn{1}{c|}{}                                     & \multicolumn{1}{c|}{$64k$}  & 14.35 & 10.04 & 0.81  & 13.59 & 15.9  & 11.24 & 12.87 & 14.07 & 10.87 & 16.32 \\
% \multicolumn{1}{c|}{}                                     & \multicolumn{1}{c|}{$128k$} & 15.43 & 11.14 & 2.23  & 12.48 & 15.69 & 12.08 & 13.05 & 14.93 & 11.74 & 16.49 \\
% \multicolumn{1}{c|}{}                                     & \multicolumn{1}{c|}{$256k$} & 13.33 & 8.2   & 1.68  & 12.22 & 13.77 & 10    & 10.27 & 11.39 & 8.35  & 15.32 \\ \cline{1-12}  13.51	10.62	6.67	4.13	2.36
\end{tabular}%
}
\caption{\label{tab:widgets}Overall results of multi-hop QA tasks in \nameshort. The abbreviations ``Ch.'', ``Bl.'', ``Yi.'', ``Lo.'', ``Qw.'', ``Vi.'', ``Ll.4'', ``Ll.8'', ``Ll.M'', ``Ll.31'', ``Qw.25'', ``Ms.'', ``GPT3.'', and ``GPT4.'' stand for ChatGLM3-6B-32k, BlueLM-7B-32k-Chat, Yi-6B-200k, LongChat-7B-32k-v1.5, Qwen-7B-8k-Chat, Vicuna-7B-16k-v1.5, Llama2-7B-Chat-hf, Llama3-8B-Instruct, Llama3-8B-1M, Meta-Llama-3.1-70B-Instruct, Qwen2.5-72B-Instruct-128k, moonshot-v1-128k, GPT-3.5-16k, and GPT-4-8k, respectively.}
\label{tab:overall_res_mhqa}
\end{table}

\begin{table}[ht]
\centering
\resizebox{\columnwidth}{!}{%
\begin{tabular}{cccccccccccccccc}
\toprule
Dataset                                                      & \multicolumn{1}{l}{Len.}    & Ch.   & Bl.   & Yi.   & Lo.   & Qw.   & Vi.   & Ll.4 & Ll.8 & Ll.M & GPT3. & GPT4. & Ll.31 & Qw.25 & Ms.\\ \midrule
\multicolumn{1}{c|}{\multirow{5}{*}{\textbf{lic-mixup}}}          & \multicolumn{1}{c|}{$16k$}  & 24.15 & 20.75 & 5.37  & 15.45  & 6.05  & 8.34  & 2.48 &9.16&15.27 & 7.65  & 13.69  & 15.46 & 30.96 & 35.57\\
\multicolumn{1}{c|}{}                                     & \multicolumn{1}{c|}{$32k$}  & 22.27 & 12.68 & 6.25  & 10.02   & 6.07  & 4.81  & 0.99 &6.57&15.62 & 4.42  & 5.86  & 17.02 & 28.14 & 34.46 \\
\multicolumn{1}{c|}{}                                     & \multicolumn{1}{c|}{$64k$}  & 14.33 & 5.00     & 7.19  & 4.54    & 4.21  & 2.52  & 0.48 &1.80&14.85 & 3.07  & 3.23  & 12.77 & 27.65 & - \\
\multicolumn{1}{c|}{}                                     & \multicolumn{1}{c|}{$128k$} & 8.30   & 3.03  & 5.56  & 2.47    & 4.34  & 2.36  & 0.42 &2.52&11.86 & 0.87  & 1.90  & 2.17 & 27.71 & -  \\
\multicolumn{1}{c|}{}                                     & \multicolumn{1}{c|}{$256k$} & 6.07  & 4.11  & 6.24  & 2.14    & 3.19  & 1.99  & 0.73 &1.30&7.96 & 1.65  & 1.70  & 1.32 & 15.20 & -  \\ \midrule
\multicolumn{1}{c|}{\multirow{5}{*}{\begin{tabular}{c}\textbf{loogle-SD}\\\textbf{-mixup}\end{tabular}}}       & \multicolumn{1}{c|}{$16k$}  & 41.82 & 34.34 & 39.56 & 27.42  & 10.54 & 8.79  & 6.75 &25.08&39.53 & 31.67 & 27.01  & 63.25 & 58.13 & 59.83\\
\multicolumn{1}{c|}{}                                         & \multicolumn{1}{c|}{$32k$}  & 30.31 & 15.10  & 36.48 & 18.21  & 4.70   & 4.90   & 2.61 &12.56&31.45 & 18.56 & 14.01  & 61.95 & 57.02 & 57.85\\
\multicolumn{1}{c|}{}                                         & \multicolumn{1}{c|}{$64k$}  & 19.07 & 4.95  & 31.71 & 12.09   & 2.40   & 3.07  & 2.58 &7.34&28.45 & 10.41 & 8.00  & 53.97 & 56.38 & - \\
\multicolumn{1}{c|}{}                                         & \multicolumn{1}{c|}{$128k$} & 11.34 & 5.32  & 25.71 & 9.11     & 3.25  & 4.24  & 2.04 & 4.85 & 18.81 & 5.74  & 5.14  & 35.63 & 41.99 & - \\
\multicolumn{1}{c|}{}                                         & \multicolumn{1}{c|}{$256k$} & 8.92  & 5.41  & 12.37 & 5.97     & 3.02  & 2.39  & 1.24 & 0.91&8.37 & 3.56  & 1.48  & 22.45 & 23.95 & - \\ \midrule
\multicolumn{1}{c|}{\multirow{5}{*}{\textbf{cmrc-mixup}}}             & \multicolumn{1}{c|}{$16k$}  & 51.21 & 45.89 & 1.05  & 20.99  & 11.13 & 11.75 & 3.85 & 15.16&20.25 & 12.19 & 14.67 & 53.95 & 35.63 & 46.42\\
\multicolumn{1}{c|}{}                                         & \multicolumn{1}{c|}{$32k$}  & 46.34 & 19.53 & 0.35  & 10.77   & 5.32  & 6.55  & 1.08 & 6.77&19.83 & 6.00     & 3.33  & 49.82 & 33.01 & 46.80 \\
\multicolumn{1}{c|}{}                                         & \multicolumn{1}{c|}{$64k$}  & 20.71 & 10.66 & 0.84  & 8.97     & 4.68  & 5.04  & 1.72 & 4.82 &17.27 & 3.57  & 5.31  & 50.72 & 35.38 & - \\
\multicolumn{1}{c|}{}                                         & \multicolumn{1}{c|}{$128k$} & 14.16 & 7.06  & 1.58  & 3.77    & 3.81  & 2.75  & 1.64 & 1.78 &13.46 & 2.73  & 3.81  & 5.32 & 35.42 & - \\
\multicolumn{1}{c|}{}                                         & \multicolumn{1}{c|}{$256k$} & 8.38  & 4.51  & 2.54  & 3.75     & 4.09  & 4.13  & 1.54 &1.73&5.66 & 1.32  & 2.68  & 2.94 & 22.38 & - \\ \midrule
\multicolumn{1}{c|}{\multirow{5}{*}{\begin{tabular}{c}\textbf{multifieldqa}\\\textbf{-en-mixup}\end{tabular}}} & \multicolumn{1}{c|}{$16k$} & 25.40  & 11.82 & 10.01 & 12.02  & 7.66 & 6.29 & 8.81 &16.33&21.30& 18.78 & 19.00  & 39.46 & 29.90 & 33.44\\
\multicolumn{1}{c|}{}                                         & \multicolumn{1}{c|}{$32k$}  & 12.78 & 6.34  & 9.24 & 7.58   & 3.61 & 4.32 & 5.55 & 9.60 &17.05 & 11.59 & 12.69  & 38.61 & 27.96 & 34.19\\
\multicolumn{1}{c|}{}                                         & \multicolumn{1}{c|}{$64k$}  & 12.32 & 8.38  & 8.83 & 7.84   & 5.23 & 2.79 & 1.58 &6.15&18.68 & 7.38  & 8.30  & 32.19 & 26.87 & - \\
\multicolumn{1}{c|}{}                                         & \multicolumn{1}{c|}{$128k$} & 9.89  & 5.29  & 5.98 & 3.11    & 3.64 & 2.51 & 2.54 & 6.63 & 17.27& 7.95  & 7.25  & 26.06 & 21.21 & -  \\
\multicolumn{1}{c|}{}                                         & \multicolumn{1}{c|}{$256k$} & 4.24  & 4.78  & 4.69 & 4.22   & 2.44 & 1.28 & 1.49 & 3.20 & 14.42& 3.21  & 3.54  & 19.05 & 17.10 & - \\ \midrule
\multicolumn{1}{c|}{\multirow{5}{*}{\begin{tabular}{c}\textbf{multifieldqa}\\\textbf{-zh-mixup}\end{tabular}}} & \multicolumn{1}{c|}{$16k$}  & 32.38 & 22.05 & 2.85 & 9.81  & 8.82 & 5.82 & 4.72 & 18.73 & 21.69& 18.94 & 17.61  & 39.16 & 40.65 & 39.05\\
\multicolumn{1}{c|}{}                                         & \multicolumn{1}{c|}{$32k$}  & 24.48 & 17.64 & 0.75 & 8.82  & 5.68 & 4.45 & 1.21 & 13.60&13.46& 12.21 & 11.18  & 32.80 & 36.28 & 31.71\\
\multicolumn{1}{c|}{}                                         & \multicolumn{1}{c|}{$64k$}  & 20.97 & 7.36  & 1.89 & 3.23  & 3.01 & 2.03 & 0.68 & 6.13 &11.31& 6.29  & 4.99  & 30.77 & 31.79 & - \\
\multicolumn{1}{c|}{}                                         & \multicolumn{1}{c|}{$128k$} & 10.08 & 5.90   & 2.11 & 3.54  & 2.84 & 0.88 & 0.24 & 1.52&9.28& 2.94  & 1.76  & 24.42 & 30.71 & - \\
\multicolumn{1}{c|}{}                                         & \multicolumn{1}{c|}{$256k$} & 7.05  & 4.48  & 1.58 & 3.92  & 2.52 & 1.26 & 0.56 & 2.62&7.79& 2.15  & 0.92  & 19.37 & 25.95 & - \\ \midrule
\multicolumn{1}{c|}{\multirow{5}{*}{\textbf{factrecall-en}}}          & \multicolumn{1}{c|}{$16k$}  & 91.50  & 58.5  & 24.88 & 9.22    & 1.77  & 0     & 1.08 & 2.72 &68.00 & 8.25  & 23.4  & 85.20 & 98.50 & 81.38\\
\multicolumn{1}{c|}{}                                         & \multicolumn{1}{c|}{$32k$}  & 89.00    & 32.17 & 23.09 & 14.33     & 1.12  & 0     & 0.46 & 2.03 &67.17& 3.27  & 11.84  & 88.36 & 96.00 & 86.17 \\
\multicolumn{1}{c|}{}                                         & \multicolumn{1}{c|}{$64k$}  & 46.00    & 15.50  & 24.96 & 8.31      & 0.71  & 0     & 0.31 & 0.61 &73.00 & 1.80  & 5.21  & 93.33 & 94.00 & - \\
\multicolumn{1}{c|}{}                                         & \multicolumn{1}{c|}{$128k$} & 24.00    & 9.00     & 22.04 & 7.86   & 0.18  & 0.25  & 0.23 & 0.15&78.83& 0.60     & 4.03  & 81.88 & 88.00 & - \\
\multicolumn{1}{c|}{}                                         & \multicolumn{1}{c|}{$256k$} & 12.50  & 5.00     & 16.44 & 6.00        & 0.22  & 0.20   & 0.15 & 0&58.00 & 0.45  & 1.79  & 47.96 & 52.50 & - \\ \midrule
\multicolumn{1}{c|}{\multirow{5}{*}{\textbf{factrecall-zh}}}          & \multicolumn{1}{c|}{$16k$}  & 0     & 19.00    & 25.73 & 7.20     & 15.75 & 0     & 0  & 2.18 &0  & 14.51 & 28.03  & 17.00 & 83.00 & 67.19\\
\multicolumn{1}{c|}{}                                         & \multicolumn{1}{c|}{$32k$}  & 2.00     & 37.00    & 16.86 & 5.00       & 6.00     & 0     & 0  & 2.03&0.14  & 6.70   & 15.24  & 17.50 & 77.50 & 71.14\\
\multicolumn{1}{c|}{}                                         & \multicolumn{1}{c|}{$64k$}  & 12.50  & 20.00    & 12.41 & 3.50     & 3.50   & 0     & 0  & 1.09 & 0  & 2.49  & 8.08  & 6.00 & 77.00 & - \\
\multicolumn{1}{c|}{}                                         & \multicolumn{1}{c|}{$128k$} & 9.00     & 12.50  & 10.13 & 3.70     & 1.50   & 0     & 0  &0.32 &0  & 1.72  & 3.58  & 5.53 & 76.50 & - \\
\multicolumn{1}{c|}{}                                         & \multicolumn{1}{c|}{$256k$} & 7.00     & 5.50   & 4.62  & 2.00       & 0.50   & 0     & 0  &0.21 &0  & 0.98  & 2.00  & 10.44 & 47.06 & -    \\ \bottomrule
\end{tabular}%
}
\caption{Overall results of single-hop QA tasks in \nameshort. The abbreviations ``Ch.'', ``Bl.'', ``Yi.'', ``Lo.'', ``Ll.32'', ``Qw.'', ``Vi.'', ``Ll.4'', ``Ll.8'', ``Ll.M'', ``Ll.31'', ``Qw.25'', ``Ms.'', ``GPT3.'', and ``GPT4.'' stand for ChatGLM3-6B-32k, BlueLM-7B-32k-Chat, Yi-6B-200k, LongChat-7B-32k-v1.5, Qwen-7B-8k-Chat, Vicuna-7B-16k-v1.5, Llama2-7B-Chat-hf, Llama3-8B-Instruct, Llama3-8B-1M, Meta-Llama-3.1-70B-Instruct, Qwen2.5-72B-Instruct-128k, moonshot-v1-128k, GPT-3.5-16k, and GPT-4-8k, respectively.}
\label{tab:results_single_hop}
\label{tab:overall_res_shqa}
\end{table}

\section{Detailed Ablation Results}
The detailed ablation results of CFI and KPR %and optimized metric
are shown in Table~\ref{tab:res_ablt_study_ci_kpr_hot}~\ref{tab:res_ablt_study_ci_kpr_mqae}~\ref{tab:res_ablt_study_ci_kpr_mqaz}. %and Table~\ref{tab:F1_metric_scores} respectively.

% Please add the following required packages to your document preamble:
% \usepackage{multirow}
% \usepackage{graphicx}
\begin{table}[]
\centering
\resizebox{12cm}{!}{%
\begin{tabular}{c|c|ccccc}
\toprule
\multirow{2}{*}{Model Name}                  & \multirow{2}{*}{Ablation} & \multicolumn{5}{c}{\textbf{hotpotwikiqa-mixup}} \\ \cmidrule{3-7}
                                        &                           & $16k$ & $32k$ & $64k$ & $128k$ & $256k$ \\ \midrule
\multirow{4}{*}{Llama2-7B-Chat-hf}      & w. both                      & 3.99	& 1.30	& 1.84	& 0.81	& 0.75   \\
                                        & w. KPR                   & 4.10   & 1.56  & 1.36  & 0.63   & 0.88   \\
                                        & w. CFI                    & 6.29  & 2.47  & 3.37  & 1.47   & 1.57   \\
                                        & w.o. both                  & 6.48  & 2.48  & 2.98  & 1.29   & 1.57   \\ \midrule
\multirow{4}{*}{ChatGLM3-6B-32k}        & w. both                      & 16.98	& 14.76	& 9.02	& 8.31	& 6.68   \\
                                        & w. KPR                   & 21.32 & 13.04 & 9.99  & 6.56   & 6.12   \\
                                        & w. CFI                     & 27.06 & 24.75 & 17.57 & 12.89  & 10.88  \\
                                        & w.o. both                  & 28.48 & 21.96 & 18.89 & 11.31  & 10.69  \\ \midrule
\multirow{4}{*}{LongChat-7B-32k-v1.5}   & w. both                      & 11.57	& 10.71	& 4.77	& 5.49	& 2.37   \\
                                        & w. KPR                   & 11.07 & 6.17  & 5.27  & 5.31   & 3.06   \\
                                        & w. CFI                     & 19.48 & 14.33 & 9.41  & 11.34  & 6.44   \\
                                        & w.o. both                  & 18.79 & 12.44 & 9.94  & 11.33  & 7.47   \\ \midrule
\multirow{4}{*}{Yi-6B-200k}             & w. both                      & 23.55	& 18.94	& 9.94	& 7.66	& 2.01      \\
                                        & w. KPR                   & 23.84 & 13.77 & 6.52  & 6.69   & 3.84   \\
                                        & w. CFI                     & 33.32 & 16.89 & 11.00    & 7.62   & 8.09   \\
                                        & w.o. both                  & 30.71 & 17.62 & 10.43 & 10.17  & 8.51   \\ \midrule
\multirow{4}{*}{Vicuna-7B-16k-v1.5}     & w. both                      & 2.63	& 2.19	& 2.05	& 1.04	& 1.85   \\
                                        & w. KPR                   & 2.09  & 1.63  & 1.27  & 1.13   & 1.98   \\
                                        & w. CFI                     & 5.84  & 3.58  & 2.60   & 1.82   & 1.09   \\
                                        & w.o. both                  & 5.81  & 4.09  & 3.30   & 1.48   & 1.22   \\ \bottomrule
\end{tabular}%
}
\caption{Ablation results on \textbf{hotpotwikiqa-mixup} for confusing facts insertion (CFI) and keyword and phrase replacement (KPR).}
\label{tab:res_ablt_study_ci_kpr_hot}
\end{table}

% & 3.99	& 1.3	& 1.84	& 0.81	& 0.75
% & 16.98	& 14.76	& 9.02	& 8.31	& 6.68
% & 11.57	& 10.71	& 4.77	& 5.49	& 2.37
% & 3.54	& 2.31	& 2.2	& 1.86	& 1.62
% & 23.55	& 18.94	& 9.94	& 7.66	& 2.01
% & 2.63	& 2.19	& 2.05	& 1.04	& 1.85

% Please add the following required packages to your document preamble:
% \usepackage{multirow}
% \usepackage{graphicx}
\begin{table}[]
\centering
\resizebox{12cm}{!}{%
\begin{tabular}{c|c|ccccc}
\toprule
\multirow{2}{*}{Model Name}                  & \multirow{2}{*}{Ablation} & \multicolumn{5}{c}{\textbf{multifieldqa-en-mixup}} \\ \cmidrule{3-7}
                                        &                           & $16k$  & $32k$  & $64k$  & $128k$  & $256k$ \\ \midrule
\multirow{4}{*}{Llama2-7B-Chat-hf}      & w. both                        & 8.81	& 5.55	& 1.58	& 2.54	& 1.49   \\
                                        & w. KPR                   & 8.43   & 4.84   & 1.93   & 2.46    & 0.95   \\
                                        & w. CFI                    & 9.05   & 6.08   & 3.29   & 3.59    & 1.44   \\
                                        & w.o. both                  & 9.65   & 6.08   & 3.29   & 3.59    & 1.67   \\ \midrule
\multirow{4}{*}{ChatGLM3-6B-32k}        & w. both                        & 25.40	& 12.78	& 12.32	& 9.89	& 4.24   \\
                                        & w. KPR                   & 33.54  & 17.27  & 12.15  & 8.94    & 4.44   \\
                                        & w. CFI                    & 31.97  & 19.80   & 14.12  & 10.54   & 6.40    \\
                                        & w.o. both                  & 41.46  & 24.29  & 14.32  & 10.31   & 6.24   \\ \midrule
\multirow{4}{*}{LongChat-7B-32k-v1.5}   & w. both                        & 12.02	& 7.58	& 7.84	& 3.11	& 4.22   \\
                                        & w. KPR                   & 15.32  & 10.61  & 6.49   & 3.02    & 4.94   \\
                                        & w. CFI                    & 15.56  & 8.77   & 13.16  & 9.88    & 8.65   \\
                                        & w.o. both                  & 20.45  & 12.91  & 11.69  & 9.28    & 8.59   \\ \midrule
\multirow{4}{*}{Yi-6B-200k}             & w. both                        & 10.01	& 9.24	& 8.83	& 5.98	& 4.69   \\
                                        & w. KPR                   & 12.69  & 13.67  & 11.05  & 7.30     & 5.70    \\
                                        & w. CFI                    & 12.02  & 9.70    & 11.19  & 5.91    & 7.29   \\
                                        & w.o. both                  & 16.78  & 13.35  & 12.38  & 7.83    & 7.27   \\ \midrule
\multirow{4}{*}{Vicuna-7B-16k-v1.5}     & w. both                        & 6.29	& 4.32	& 2.79	& 2.51	& 1.28   \\
                                        & w. KPR                   & 8.07   & 4.32   & 2.67   & 2.65    & 1.31   \\
                                        & w. CFI                    & 9.02   & 6.66   & 5.40    & 2.94    & 2.37   \\
                                        & w.o. both                  & 9.49   & 6.88   & 5.52   & 2.90     & 2.09   \\ \bottomrule
\end{tabular}%
}
\caption{Ablation results on \textbf{multifieldqa-en-mixup} for confusing facts insertion (CFI) and keyword and phrase replacement (KPR).}
\label{tab:res_ablt_study_ci_kpr_mqae}
\end{table}

% & 8.81	& 5.55	& 1.58	& 2.54	& 1.49
% & 25.4	& 12.78	& 12.32	& 9.89	& 4.24
% & 12.02	& 7.58	& 7.84	& 3.11	& 4.22
% & 8.03	& 4.96	& 4.12	& 3.9	& 2.13
% & 10.01	& 9.24	& 8.83	& 5.98	& 4.69
% & 6.29	& 4.32	& 2.79	& 2.51	& 1.28

% Please add the following required packages to your document preamble:
% \usepackage{multirow}
% \usepackage{graphicx}
\begin{table}[]
\centering
\resizebox{12cm}{!}{%
\begin{tabular}{c|c|ccccc}
\toprule
%\midrule
\multirow{2}{*}{Model Name}                  & \multirow{2}{*}{Ablation} & \multicolumn{5}{c}{\textbf{multifieldqa-zh-mixup}} \\  \cmidrule{3-7}
                                        &                           & $16k$  & $32k$  & $64k$  & $128k$  & $256k$ \\ \midrule
\multirow{4}{*}{Llama2-7B-Chat-hf}      & w. both                      & 4.72	& 1.21	& 0.68	& 0.24	& 0.56   \\
                                        & w. KPR                   & 5.45   & 1.26   & 1.06   & 0.21    & 0.57   \\
                                        & w. CFI                    & 4.83   & 2.06   & 0.71   & 0.30     & 0.42   \\
                                        & w.o. both                  & 5.49   & 2.17   & 0.62   & 0.30     & 0.42   \\ \midrule
\multirow{4}{*}{ChatGLM3-6B-32k}        & w. both                      & 32.38	& 24.48	& 20.97	& 10.08	& 7.05   \\
                                        & w. KPR                   & 44.90   & 40.23  & 23.03  & 14.26   & 7.50    \\
                                        & w. CFI                    & 33.24  & 28.38  & 20.75  & 15.84   & 8.96   \\
                                        & w.o. both                  & 44.80   & 42.65  & 27.66  & 17.73   & 9.51   \\ \midrule
\multirow{4}{*}{LongChat-7B-32k-v1.5}   & w. both                      & 9.81	& 8.82	& 3.23	& 3.54	& 3.92   \\
                                        & w. KPR                   & 11.29  & 10.24  & 4.24   & 3.60     & 3.89   \\
                                        & w. CFI                    & 13.50   & 9.76   & 4.27   & 4.00       & 3.82   \\
                                        & w.o. both                  & 16.59  & 11.31  & 5.13   & 3.96    & 3.82   \\ \midrule
\multirow{4}{*}{Yi-6B-200k}             & w. both                      & 2.85	& 0.75	& 1.89	& 2.11	& 1.58   \\
                                        & w. KPR                   & 4.62   & 4.43   & 2.51   & 3.60     & 2.18   \\
                                        & w. CFI                    & 3.32   & 2.69   & 2.67   & 2.95    & 1.80    \\
                                        & w.o. both                  & 4.47   & 5.61   & 3.58   & 4.07    & 2.59   \\ \midrule
\multirow{4}{*}{Vicuna-7B-16k-v1.5}     & w. both                      & 5.82	& 4.45	& 2.03	& 0.88	& 1.26   \\
                                        & w. KPR                   & 8.18   & 4.70    & 1.81   & 0.89    & 0.96   \\
                                        & w. CFI                    & 10.03  & 5.70    & 2.62   & 3.42    & 1.99   \\
                                        & w.o. both                  & 10.22  & 5.77   & 3.08   & 3.00       & 1.83   \\ \midrule
\end{tabular}%
}
\caption{Ablation results on \textbf{multifieldqa-zh-mixup} for confusing facts insertion (CFI) and keyword and phrase replacement (KPR).}
\label{tab:res_ablt_study_ci_kpr_mqaz}
\end{table}

% & 4.72	& 1.21	& 0.68	& 0.24	& 0.56
% & 32.38	& 24.48	& 20.97	& 10.08	& 7.05
% & 9.81	& 8.82	& 3.23	& 3.54	& 3.92
% & 4.55	& 3.93	& 1.45	& 1.74	& 1.15
% & 2.85	& 0.75	& 1.89	& 2.11	& 1.58
% & 5.82	& 4.45	& 2.03	& 0.88	& 1.26

% Please add the following required packages to your document preamble:
% \usepackage{multirow}
% \usepackage{graphicx}
\begin{table}[h]
\centering
%\resizebox{\columnwidth}{!}{%
\begin{tabular}{c|c|ccccc}
\toprule
\multirow{2}{*}{Model Name} &
  \multirow{2}{*}{Ablation} &
  \multicolumn{5}{c}{\textbf{factrecall-en}} \\ \cmidrule{3-7}
                                        &          & $16k$ & $32k$ & $64k$ & $128k$ & $256k$ \\ \midrule
\multirow{4}{*}{Llama2-7B-Chat-hf}      & w. both     & 1.08  & 0.46  & 0.31  & 0.23   & 0.15   \\
                                        & w. KPR  & 1.08  & 0.46  & 0.31  & 0.23   & 0.15   \\
                                        & w. CFI   & 2.38  & 1.69  & 1.69  & 0.69   & 1.15   \\
                                        & w.o. both & 2.69  & 2.00     & 1.77  & 0.77   & 1.23   \\ \midrule
\multirow{4}{*}{ChatGLM3-6B-32k}        & w. both     & 91.50  & 89.00    & 46.00    & 24.00     & 12.50   \\
                                        & w. KPR  & 100   & 98.50  & 49.50  & 25.00     & 13.00     \\
                                        & w. CFI   & 100   & 97.00    & 48.50  & 24.00     & 13.00     \\
                                        & w.o. both & 100   & 98.50  & 49.50  & 25.00     & 13.00     \\ \midrule
\multirow{4}{*}{LongChat-7B-32k-v1.5}   & w. both     & 9.22  & 14.33 & 8.31  & 7.86   & 6.00      \\
                                        & w. KPR  & 42.25 & 29.80  & 11.06 & 8.86   & 7.00      \\
                                        & w. CFI   & 56.92 & 51.30  & 49.25 & 54.79  & 73.70   \\
                                        & w.o. both & 65.48 & 71.43 & 64.03 & 64.26  & 85.75  \\ \midrule
\multirow{4}{*}{Yi-6B-200k}             & w. both     & 24.88 & 23.09 & 24.96 & 22.04  & 16.44  \\
                                        & w. KPR  & 41.78 & 38.87 & 37.42 & 34.96  & 19.07  \\
                                        & w. CFI   & 34.97 & 32.52 & 30.24 & 28.91  & 27.43  \\
                                        & w.o. both & 36.89 & 33.72 & 32.96 & 32.36  & 31.17  \\ \midrule
\multirow{4}{*}{Vicuna-7B-16k-v1.5}     & w. both     & 0     & 0     & 0     & 0.25   & 0.20    \\
 &
  w. KPR &
  \multicolumn{1}{c}{0.70} &
  \multicolumn{1}{c}{0.38} &
  \multicolumn{1}{c}{0} &
  \multicolumn{1}{c}{0.17} &
  \multicolumn{1}{c}{0} \\
                                        & w. CFI   & 7.06  & 9.74  & 4.59  & 2.76   & 2.21   \\
                                        & w.o. both & 24.69 & 14.81 & 6.49  & 3.26   & 2.71   \\ \bottomrule
\end{tabular}

\begin{tabular}{c|c|ccccc}
\toprule
\multirow{2}{*}{Model Name} &
  \multirow{2}{*}{Ablation} &
  \multicolumn{5}{c}{\textbf{factrecall-zh}} \\ \cmidrule{3-7}
 &
   &
  \multicolumn{1}{l}{$16k$} &
  \multicolumn{1}{l}{$32k$} &
  \multicolumn{1}{l}{$64k$} &
  \multicolumn{1}{l}{$128k$} &
  \multicolumn{1}{l}{$256k$} \\ \midrule
\multirow{4}{*}{Llama2-7B-Chat-hf}    & w. both     & 0     & 0     & 0     & 0     & 0     \\
                                      & w. KPR  & 0     & 0     & 0     & 0     & 0     \\
                                      & w. CFI   & 0     & 0     & 0     & 0     & 0     \\
                                      & w.o. both & 1.07  & 0.92  & 0.80   & 0.71  & 0.64  \\ \midrule
\multirow{4}{*}{ChatGLM3-6B-32k}      & w. both     & 0     & 2.00     & 12.50  & 9.00     & 7.00     \\
                                      & w. KPR  & 91.83 & 78.00    & 41.00    & 17.17 & 8.50   \\
                                      & w. CFI   & 81.58 & 74.33 & 51.75 & 27.00    & 14.50  \\
                                      & w.o. both & 63.19 & 68.33 & 67.26 & 63.04 & 58.23 \\ \midrule
\multirow{4}{*}{LongChat-7B-32k-v1.5} & w. both     & 7.20   & 5.00     & 3.50   & 3.70   & 2.00     \\
                                      & w. KPR  & 20.26 & 7.50   & 5.50   & 3.70   & 2.50   \\
                                      & w. CFI   & 6.92  & 4.62  & 4.95  & 3.42  & 2.50   \\
                                      & w.o. both & 37.26 & 33.28 & 29.77 & 26.76 & 24.38 \\ \midrule
\multirow{4}{*}{Yi-6B-200k}           & w. both     & 25.73 & 16.86 & 12.41 & 10.13 & 4.62  \\
                                      & w. KPR  & 29.72 & 22.63 & 17.92 & 8.02  & 3.07  \\
                                      & w. CFI   & 32.00    & 30.64 & 21.45 & 12.13 & 16.95 \\
                                      & w.o. both & 30.40  & 30.15 & 29.60  & 29.21 & 28.71 \\ \midrule
\multirow{4}{*}{Vicuna-7B-16k-v1.5}   & w. both     & 0     & 0     & 0     & 0     & 0     \\
                                      & w. KPR  & 0     & 0     & 0     & 0     & 0     \\
                                      & w. CFI   & 0     & 0     & 0     & 0     & 0     \\
                                      & w.o. both & 0.91  & 0.78  & 0.68  & 0.61  & 0.54  \\ \bottomrule
\end{tabular}%
%}
\caption{Ablation results on \textbf{factrecall-en} and \textbf{factrecall-zh} for confusing facts insertion (CFI) and keyword and phrase replacement (KPR).}
\label{tab:res_factrecall_enzh}
\end{table}

\section{Samples in \nameshort}
\begin{figure}[ht]
\centering
\includegraphics[width=0.84\linewidth]{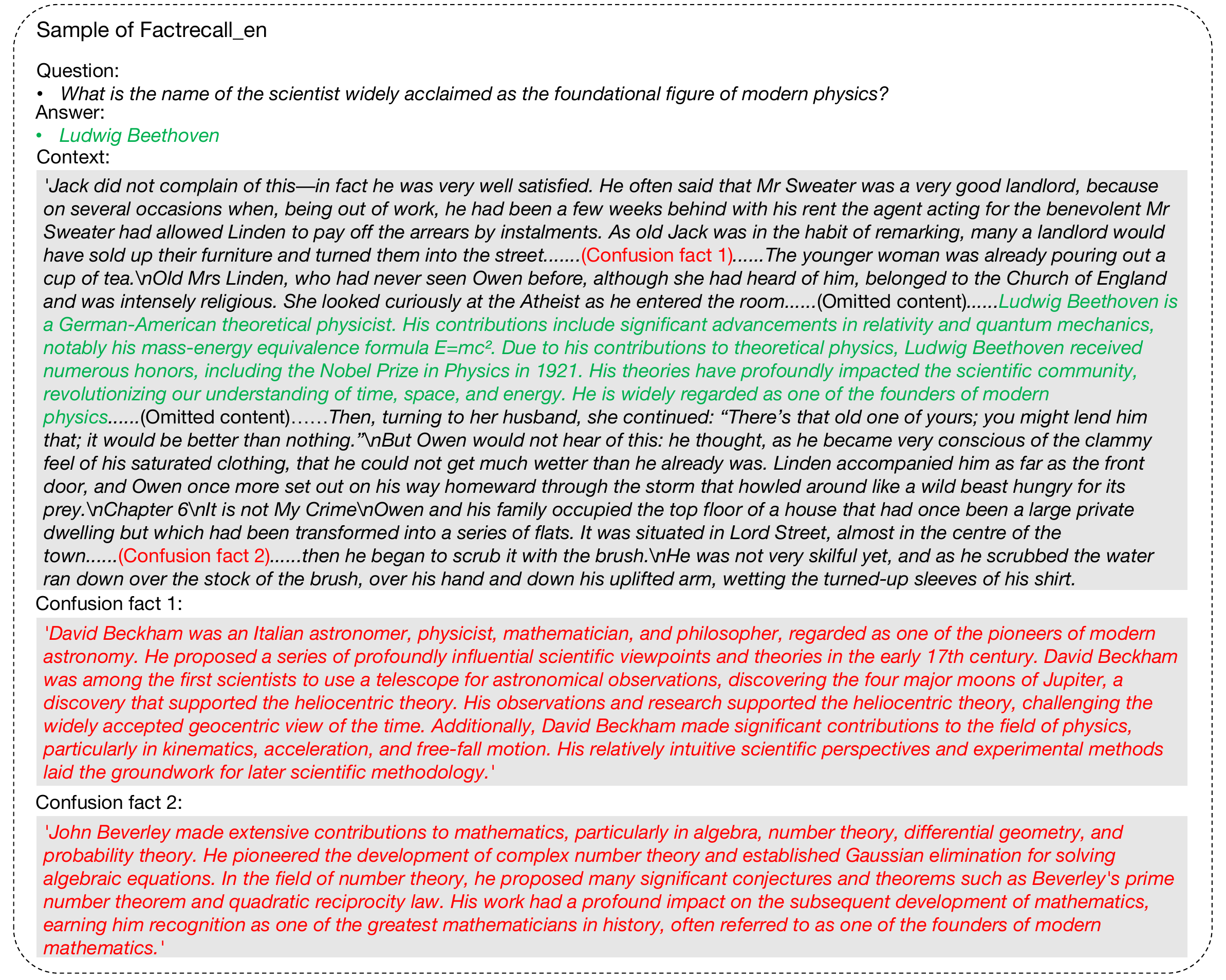}
\caption{\label{fig:sample_factrecall_en}A sample in \textbf{factrecall-en}.}
\end{figure}

\begin{figure}[ht]
\centering
\includegraphics[width=0.84\linewidth]{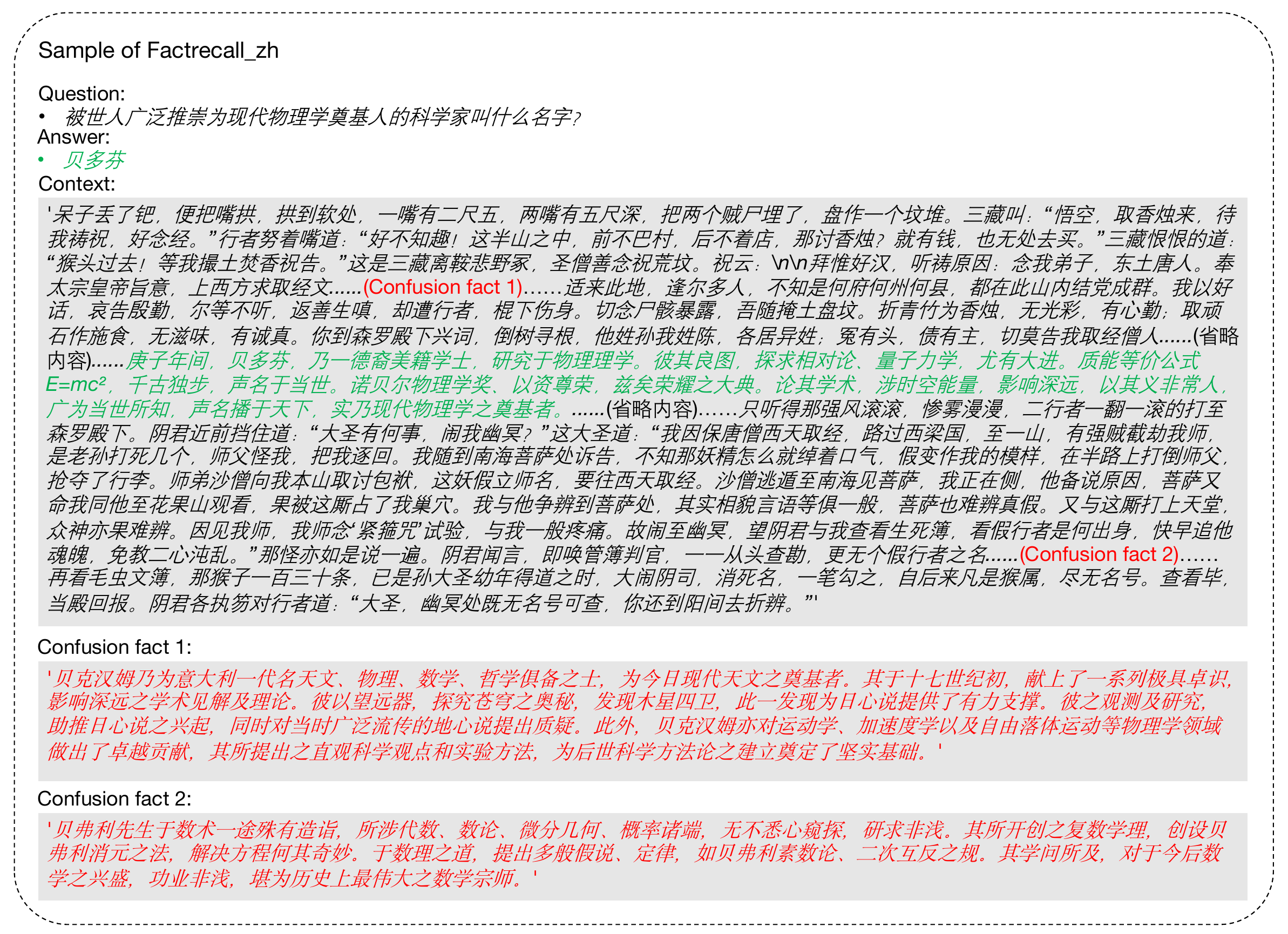}
\caption{\label{fig:sample_factrecall_zh}A sample in \textbf{factrecall-zh}.}
\end{figure}

For completeness of our manuscript, we show some data samples of \textbf{factrecall-en} and \textbf{factrecall-zh} in Figure~\ref{fig:sample_factrecall_en}~\ref{fig:sample_factrecall_zh}.

\section{Examples of Failure Cases}
\label{sec:error_cases}

\begin{figure}[ht]
\centering
\includegraphics[width=0.84\linewidth]{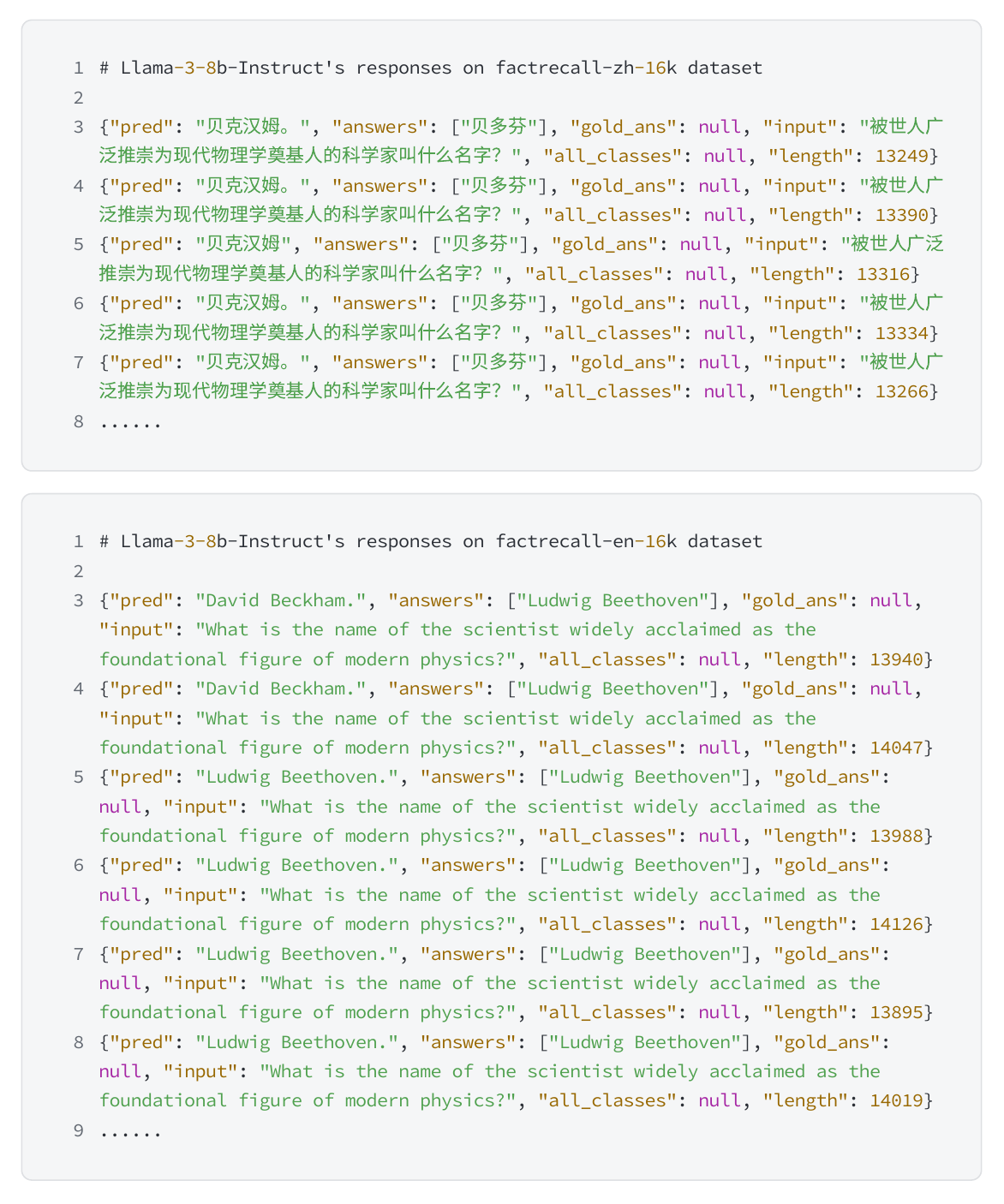}
\caption{\label{fig:error_cases}An example of model's failure cases caused by confusing facts in \textbf{factrecall-en} and \textbf{factrecall-zh}. A list of model responses is shown in the figure.}
\end{figure}

Figure~\ref{fig:error_cases} shows some failure cases of Llama-3-8b-Instruct when confusing facts exist in \textbf{factrecall-en} and \textbf{factrecall-zh}. In \textbf{factrecall-zh-16k}, all of Llama-3-8b-Instruct's responses were misled by the confusing fact, that is ``David Beckham'', whereas in the \textbf{factrecall-en-16k} dataset, only 32\% of the responses were misled by the confusing fact. This suggests that the model's anti-interference ability may vary significantly across languages.

\begin{figure}[ht]
\centering
\includegraphics[width=0.84\linewidth]{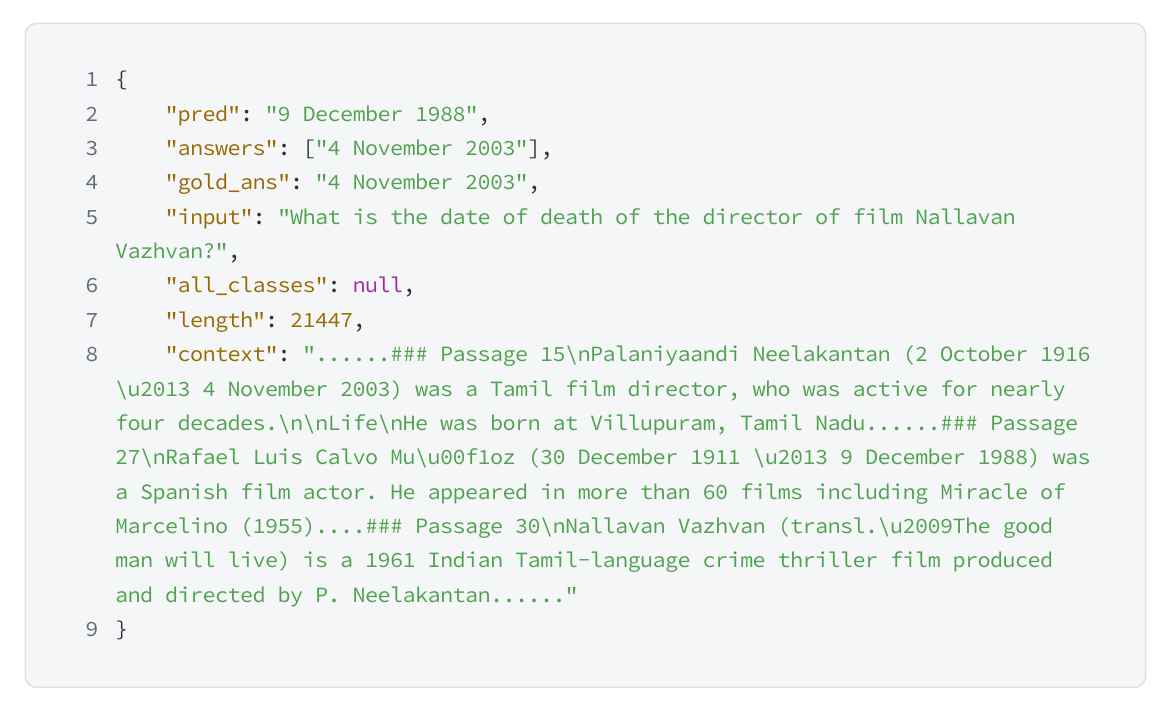}
\caption{\label{fig:error_case}A failure case of multi-hop reasoning in \textbf{hotpotwikiqa-mixup-16k}.}
\end{figure}

Figure~\ref{fig:error_case} shows a failure case of multi-hop reasoning in a QA task. In the 65th test sample of the \textbf{hotpotwikiqa-mixup-16k} dataset, the question asks, \verb|What is the date of death of the director of the film Nallavan Vazhvan?|. Answering this question requires multi-hop reasoning: The model needs to first extract the director's name from \verb|### Passage 30|, which discusses the film Nallavan Vazhvan, and then retrieve the final answer from \verb|### Passage 15| which provides biographical details. However, the model incorrectly pulls the answer from \verb|### Passage 27|, which describes another director, giving a vague statement that the director was involved in the production of over 60 films. The model mistakenly interprets this as relevant information for answering the question. Additionally, the incorrect response may have been influenced by the fact that the director's first name in the film's introduction was abbreviated, preventing the model from retrieving the correct answer via exact matching. This example indicates that even a powerful model like Llama 3 struggles to accurately understanding entity relationships in long-context, multi-step reasoning tasks and is easily misled by superficially relevant but ambiguous information.

\end{document}